\documentclass[preprint,12pt]{elsarticle}

\usepackage{times}
\usepackage{soul}
\usepackage{url}
\usepackage[hidelinks]{hyperref}
\hypersetup{
    colorlinks=true, % 启用彩色链接
    linkcolor=blue,  % 定义链接的颜色为蓝色
    citecolor=blue,  % 定义引用的颜色为蓝色
    filecolor=blue,  % 定义文件链接的颜色为蓝色
    urlcolor=blue    % 定义URL的颜色为蓝色
}
\usepackage[utf8]{inputenc}
\usepackage[small]{caption}
\usepackage{graphicx}
\usepackage{amsmath}
\usepackage{amsthm}
\usepackage{amssymb}
\usepackage{booktabs}
\usepackage{algorithm}
\usepackage{algorithmic}
\usepackage[switch]{lineno}
\usepackage{multirow}
\usepackage{float}
\usepackage{subfigure}
\usepackage[dvipsnames]{xcolor}
\usepackage{tikz}
\usepackage{xcolor}

\nolinenumbers
\usepackage{xspace}
\makeatletter
\DeclareRobustCommand\onedot{\futurelet\@let@token\@onedot}
\def\@onedot{\ifx\@let@token.\else.\null\fi\xspace}
 
\def\ie{\emph{i.e}\onedot}

\makeatother

\usepackage{color}
\definecolor{mygray}{gray}{0.6}
\definecolor{dg}{rgb}{0,0.694,0.298}
\definecolor{purple}{rgb}{0.4,0.176,0.569}
\definecolor{pink}{cmyk}{0, 0.7808, 0.4429, 0.1412}

\journal{Neuron Networks}

\begin{document}

\begin{frontmatter}

\title{Using a single actor to output personalized policy for different intersections}

\author[label1]{Kailing Zhou}
\author[label1]{Chengwei Zhang\corref{cor1}}
\author[label1]{Furui Zhan\corref{cor2}}
\author[label1]{Wanting Liu}
\author[label1]{Yihong Li}
\affiliation[label1]{organization={Dalian Maritime University},
            addressline={1 Linghai Road}, 
            city={Dalian},
            postcode={116026}, 
            state={Liaoning},
            country={China}}

\cortext[cor1]{corresponding author. E-mail: chenvy@dlmu.edu.cn}
\cortext[cor2]{corresponding author. E-mail: izfree@dlmu.edu.cn}

\begin{abstract}
Recently, with the development of Multi-agent reinforcement learning (MARL), adaptive traffic signal control (ATSC) has achieved satisfactory results. In traffic scenarios with multiple intersections, MARL treats each intersection as an agent and optimizes traffic signal control strategies through learning and real-time decision-making. To enhance the efficiency of training and deployment in large-scale intersection scenarios, existing work predominantly employs shared parameter methods. Considering that observation distributions of intersections might be different in real-world scenarios, shared parameter methods might lack diversity and thus lead to high generalization requirements in the shared-policy network. A typical solution is to increase the size of network parameters. However, simply increasing the scale of the network does not necessarily improve policy generalization, which is validated in our experiments. Moreover, practical traffic signal control systems must consider the deployment cost of decision devices. Accordingly, an approach that considers both the personalization of intersections and the efficiency of parameter sharing is required. To this end, we propose Hyper-Action Multi-Head Proximal Policy Optimization (HAMH-PPO), a Centralized Training with Decentralized Execution (CTDE) MARL method that utilizes a shared PPO policy network to deliver personalized policies for intersections with non-iid observation distributions. The centralized critic in HAMH-PPO uses graph attention units to calculate the graph representations of all intersections and outputs a set of value estimates with multiple output heads for each intersection. The decentralized execution actor takes the local observation history as input and output distributions of action as well as a so-called hyper-action to balance the multiple values estimated from the centralized critic to further guide the updating of TSC policies. The combination of hyper-action and multi-head values enables multiple agents to share a single actor-critic while achieving personalized policies. The effectiveness of HAMH-PPO is validated through extensive experiments on real-world and synthetic road network traffic.
\end{abstract}


\end{frontmatter}

\section{Introduction}
Multi-Agent Reinforcement Learning (MARL) focuses on collaboratively training a group of agents to solve specific tasks in shared environments. Recent research primarily concentrated on optimizing feature extraction techniques~\cite{ge2021multi, DBLP:journals/access/QiSL24}, refining agent modeling~\cite{DBLP:journals/corr/abs-2112-02336}, and designing hierarchical cooperative strategies~\cite{xu2021hierarchically,ABDOOS2021114580,shen2023heterogeneous}. However, a relatively overlooked key issue is how to effectively extend multi-agent algorithms to scenarios involving a large number of agents, especially in complex Traffic Signal Control (TSC) applications. Statistical data reveals that up to 96\% of related studies only involve simulation environments with fewer than 100 intersections~\cite{liu2023gplight}.
A critical yet underexplored issue in the domain of multi-agent systems pertains to the scalable extension of algorithms to accommodate a substantial number of agents, particularly within the intricate context of Traffic Signal Control. 

\begin{figure}[!h]
    \centering
    \subfigure[$1\times3$ road network structure and traffic flow distribution.]{
    \label{fig:1_3_road}
    \includegraphics[width=14cm]{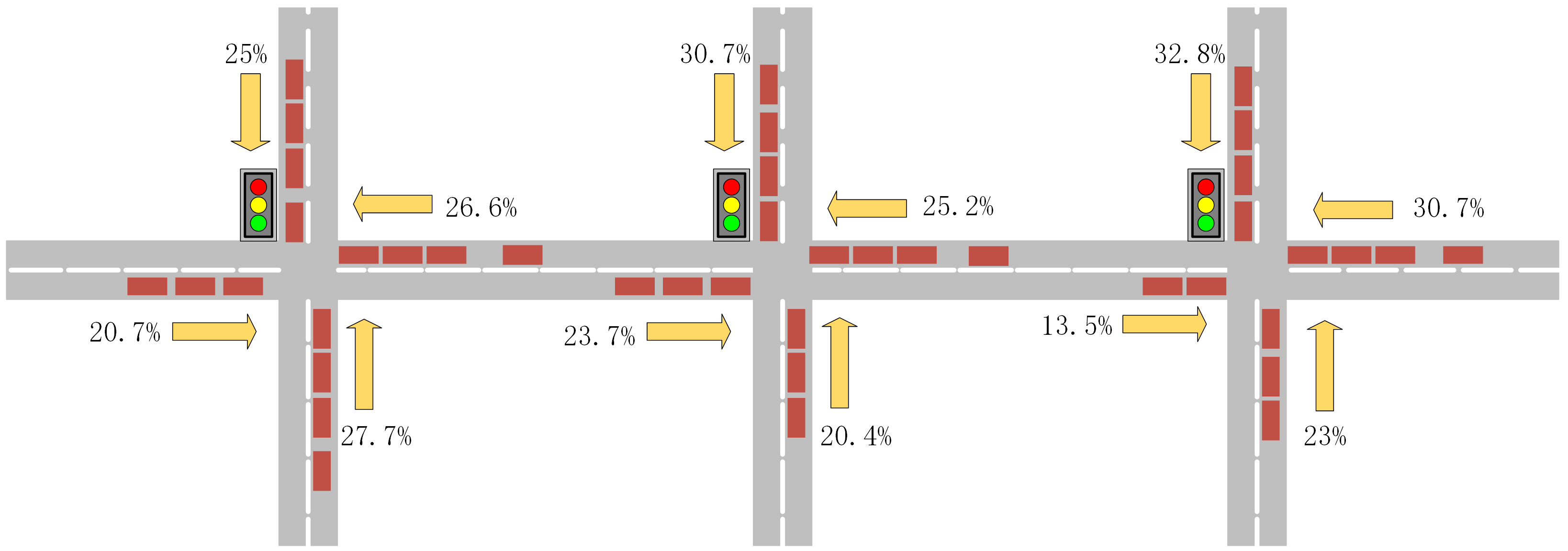}}
    \subfigure[The distribution of observations for the three intersections, processed through Principal Component Analysis (PCA) for dimensionality reduction.]{
    \label{fig:1_3_Observed_distribution}
    \includegraphics[width=6cm]{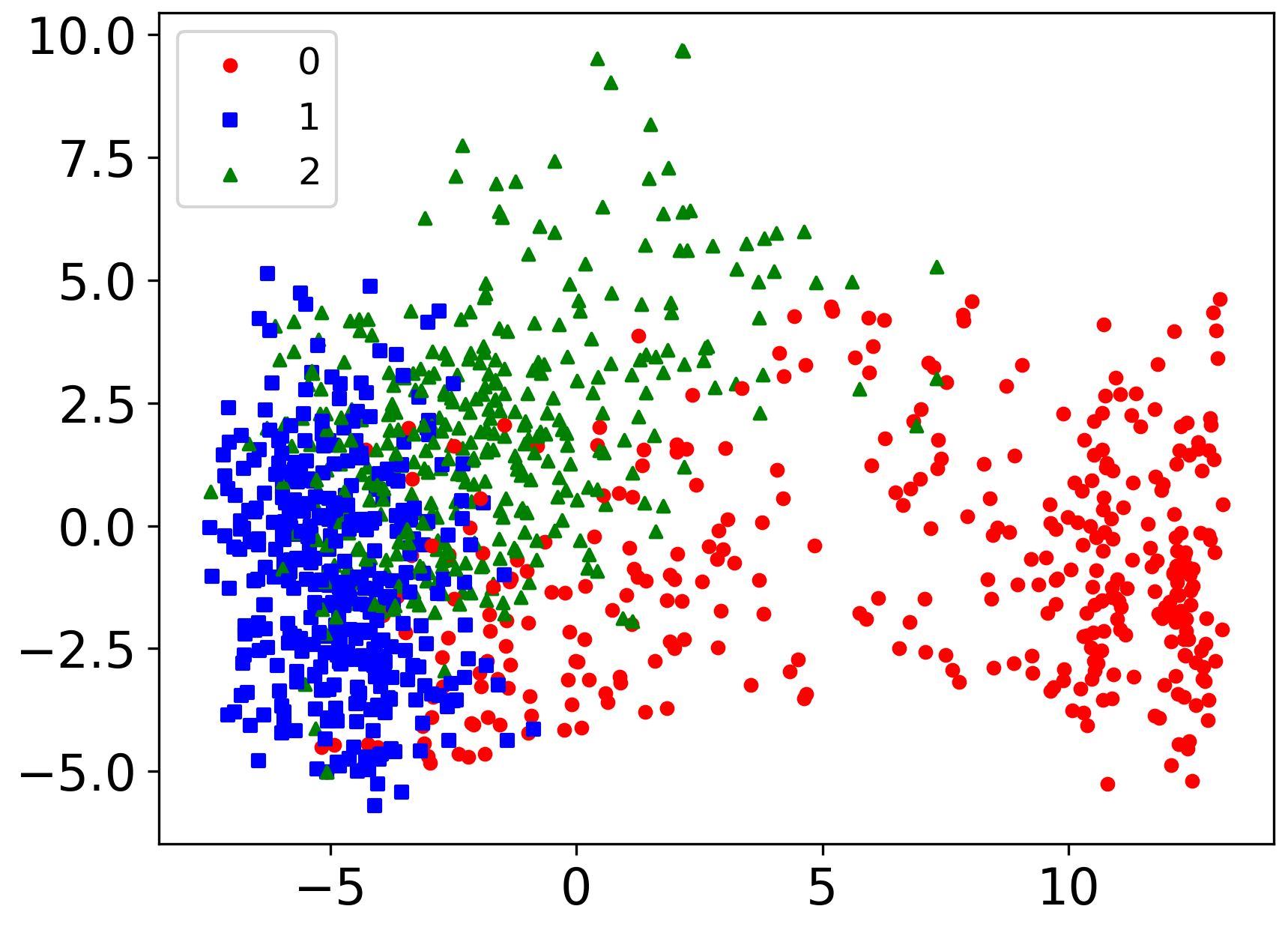}}
    \hfill
    \subfigure[Learning curves (measured by average travel time) and stacked chart of training duration (minutes) under parameter sharing(\textit{PPO-share}) and non-parameter sharing (\textit{PPO-non share}) methods.
    ]{
    \label{fig:1_3_Travel_Training_new}
    \includegraphics[width=7cm]{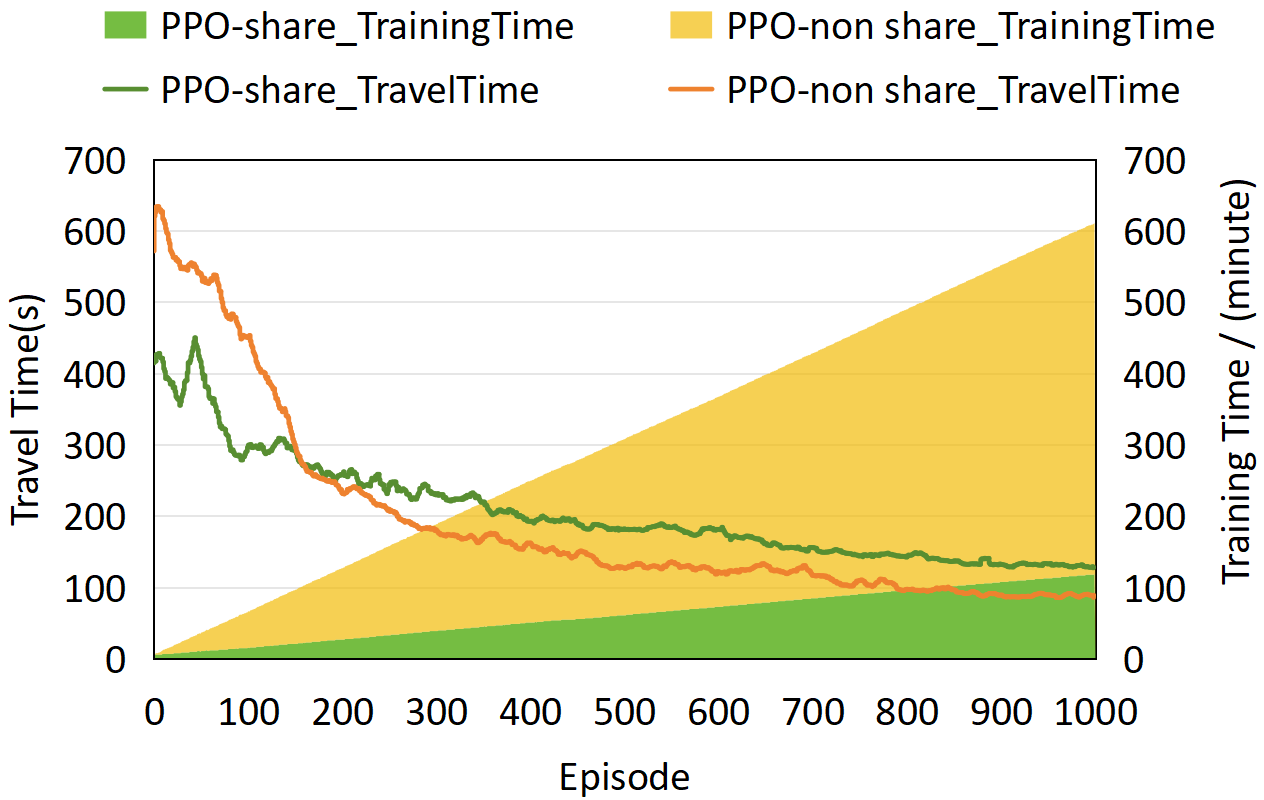}}
    \caption{The traffic flow distribution and the observed distribution of a 1 $\times$ 3 network, as well as the learning curve for parameter sharing and non-parameter sharing methods (plotted based on 5 testing runs, with smoothing applied every 9 episodes.). This means that the personalization of the intersection needs to be considered.}
    \label{fig:enter-label}
\end{figure}

To facilitate the training of multi-agent systems in large-scale traffic signal control, a common technique is parameter sharing, where network parameters are shared among agents~\cite{christianos2021scaling}. In many cases, this policy is highly effective, and algorithms such as Colight~\cite{wei2019colight} and MPLight~\cite{chen2020toward} have achieved remarkable results with parameter sharing on large-scale road networks over 2,000 intersections. This is mainly attributed to the existence of similar environmental characteristics between many intersections, allowing the agent to find similar observations and reward functions. This similarity allows agents to be uniformly represented across the neural network layers, thus speeding up the training process. However, this representation method also loses a part of the individual-specific information. Although parameter sharing can speed up training, it ignores the personalized characteristics of each intersection. In real scenarios, the difference in traffic environment is often very significant~\cite{zhang2022neighborhood}. For example, there is high traffic near schools and hospitals, while relatively few vehicles are at remote intersections. In our experiments, we observe that the simple parameter-sharing policy becomes ineffective when the traffic environment becomes complex and irregular. This indicates that it is difficult to form a unified hidden representation in a complex environment, and it is unreasonable to consider the state features of all intersections indiscriminately.
To solve this problem, some approaches~\cite{zhang2022neighborhood,chu2019multi,liu2023contrastive} try to achieve personalization by making distinctions among agents. However, this approach goes against the original purpose of large-scale problems, since that it is impractical to design unique parameters for each agent in large-scale networks. Recent work attempts to increase the diversity of agents by grouping intersections and sharing strategies within groups. For example, FMA2C~\cite{ma2020feudal} divides the traffic network into fixed areas, where different managers are used to control areas and each intersection is controlled by a worker. GPLight~\cite{liu2023gplight} dynamically groups agents based on extracted environment features and employs two loss functions to maintain a learnable dynamic clustering. Alternatively, GoMARL~\cite{zang2024automatic} proposed a sharing mechanism that specialized policy parameters by value decomposition. This selective sharing approach can accommodate agents' diversity while retaining parameter sharing's advantages. However, it relies on the capability of agent partition and does not take into account the specific preferences of each agent. Therefore, future research should further explore how to better utilize each agent's personalized characteristics while maintaining the system's efficiency.

The above demo shows an ATSC task where three consecutive intersections on a straight road have different traffic flow distributions. The heterogeneity of traffic flow distribution can also be understood as the heterogeneity of observation distribution in reinforcement learning, which results in agents having distinct optimal policies. As shown in Figure \ref{fig:1_3_road}, the proportions of vehicles in lanes approaching intersections from four directions are different. For example, the 20.7\% indicated by the yellow arrow at the first intersection represents the percentage of vehicles passing through that ingoing lane to the total flow at that intersection. The red rectangles represent the vehicle entities. To present the heterogeneity of traffic flow distribution more intuitively, we collected observations over 360 steps in an episode and plotted the distribution of these observations at three intersections using dimensionality reduction techniques. As shown in Figure \ref{fig:1_3_Observed_distribution}, the observations of three intersections are clustered in different areas, showing significant differences. To further illustrate the necessity of intersection diversity and personalized policies, we conducted a comparative analysis between two experimental sets: one training a single agent for parameter sharing (\textit{PPO-share}), and the other training an independent agent for each intersection (\textit{PPO-non share}). Both experimental sets are based on the PPO algorithm~\cite{schulman2017proximal} and are trained with 1000 episodes. The curves in Figure \ref{fig:1_3_Travel_Training_new} show the training results of the two methods with average travel time. The experimental results show that the training time of the parameter-sharing method is only 119 minutes, which is significantly less than the 492 minutes of the non-sharing method. In addition, it can be seen that the parameter-sharing method can find faster find a relatively good policy at the early training stage. However, with the deepening of training, this method gradually shows its limitations, because it fails to fully consider the uniqueness of each intersection, and it is difficult to achieve the optimal policy. In contrast, non-parameter sharing methods, although slower in the initial phase, can ultimately achieve better results. The experimental results demonstrate the importance of considering the diversity of intersections and adopting individualized strategies in the complex traffic network environment. How to effectively combine the training efficiency and the individual requirements of the intersection will be the main consideration of our research.

Our objective is to create a cooperative joint policy training framework and model mechanism for ATSC scenarios, that facilitate efficient, robust policy learning by policy parameter sharing between multiple intersections without inheriting the disadvantage of policy sharing. The local observations with different distributions push agents to local solutions at the expense of the overall average performance between multiple intersections. A similar phenomenon in federated learning is also known as \textit{client drift}~\cite{pmlr-v119-karimireddy20a}.

To more comprehensively consider the personalized requirements under shared parameters, we propose a multi-agent reinforcement learning model for traffic signal control that balances diversity and efficiency, namely Hyper-Action Multi-Head Proximal Policy Optimization (HAMH-PPO). HAMH-PPO adopts the centralized training and decentralized execution (CTDE) paradigm, where all agents share parameters. The critic network estimates multiple value functions and hyper-action output through actor network to provide personalized policy for agents.  
Our goal is to jointly guide the learning of the actor through multiple value functions. Specifically, the centralized critic takes global observations as input to estimate the travel time at each intersection and outputs a vector containing multiple value functions. The actor network takes local observations as input to capture time-dependent feature information through a Gate Recurrent Unit (GRU)\cite{DBLP:journals/corr/ChungGCB14}, and outputs actions acting on the environment. In addition, a hyper-action is used to evaluate the importance of the value function, which takes temporal features and intersection subscripts as input and outputs personalized hyper-action. Since the number of value functions is limited, selecting only one value function for estimation may not be accurate enough. The hyper-action network can directly generate the weights of the corresponding value functions and obtain the joint value function to provide more comprehensive and accurate policy guidance so that the representation ability can be enhanced.

We evaluate the performance of the algorithm with both real and synthetic datasets, especially in a large-scale road network with 100 intersections. The experimental results show that our proposed algorithm can flexibly adjust the weight according to the dynamic characteristics of the intersection, and the learned policy can effectively deal with the complex traffic environment. To summarize, the main contributions of this work are as follows:
\begin{itemize}
    \item We analyzed and verified that there are differences in the observed distribution of intersections in TSC scenarios, and these differences have a direct impact on the performance of signal light control policies.
    \item A HAMH-PPO model is proposed, which achieves the balance between diversity and efficiency of traffic signal control through the critic network of multi-valued functions and the hyper-action mechanism. Additionally, the model can provide personalized policy services for agents with different observation distributions.
    \item We conducted experiments on six datasets, including two real road networks and two synthetic road networks, to verify the efficiency and diversity of this method. We release our source code at \href{https://github.com/ET-zkl/HAMH-PPO.git}{https://github.com/ET-zkl/HAMH-PPO.git}.
\end{itemize}

The remainder of this paper is organized as follows. Section \ref{sec:Related_Work} discusses the related works. Section \ref{sec:Preliminary} describes the basic notations of ATSC and MARL. The proposed algorithm is presented in Section \ref{sec:HAMH-PPO}, and the experiments and performance analysis are detailed in Section \ref{sec:Experiments}. Finally, we conclude the article in Section \ref{sec:Conlcusion}.

\section{Related Work}
\label{sec:Related_Work}
Traditional traffic signal control methods are mostly based on expert rules and cannot adaptively adjust signal phases based on real-time traffic flow. In recent years, MARL-based adaptive traffic signal control algorithms have shown promising results. InteliLight~\cite{wei2018intellilight} emphasizes the importance of features, where the state inputs include the queue length of each lane, the total number of vehicles at the intersection, updated waiting times, images of the positions of each vehicle at the intersection, the current agent's action, and the next action. Colight~\cite{wei2019colight} first introduces graph attention networks (GAT)~\cite{velickovic2017graph} to learn the dynamic impact of neighbor intersections and capture the spatial characteristics of intersections. PDA-TSC~\cite{fang2022monitorlight} considers fairness issues and uses Long Short Term Memory (LSTM) to predict the duration of actions. MARL-DSTAN~\cite{huang2021network} and STMARL~\cite{wang2020stmarl} respectively use graph attention and graph convolution to capture spatial features of intersections, while RNN is utilized to extract temporal dependencies. GCQN-TSC~\cite{yan2023graph} is a graph cooperation Q-learning model designed with the starting point of ecological traffic. Considering the heterogeneity of intersections, IG-RL~\cite{devailly2021ig} utilizes the flexible computational graph and inductive capability of graph convolutional networks (GCNs)~\cite{hamilton2017inductive} to obtain a set of parameters suitable for various road network controls, so that the generalization and transferability of multiple agents can be enhanced. The IHG-MA~\cite{yang2021ihg} algorithm can generate embeddings for new incoming vehicles and new traffic networks. Unlike homogeneous algorithms, the IHG-MA algorithm not only encodes the heterogeneous features of each node but also encodes heterogeneous structural (graph) information. CoevoMARL~\cite{10556581} utilizes the relationship-driven Progressive LSTM (RDP-LSTM) to dynamically evolve the learned spatial interaction network of signals over time.  
CCGN~\cite{MUKHTAR2023396} enables collaboration between intersections by combining local policy networks (LPN) and global policy networks (GPN) to achieve signal-free traffic flow control. CI-MA~\cite{YANG2023243} utilizes causal inference modeling to handle the non-stationarity of multi-agent traffic environments, resulting in effective cooperative traffic signal strategies.

In these methods, multiple agents share a learning network, which greatly improves the training efficiency of the neural network. However, they do not take into account the individualization of intersections, which may lead to some performance degradation. As a result, MA2C~\cite{chu2019multi} trains independent agents for each intersection to control traffic signals through local observation and communication between neighbors. PressLight~\cite{wei2019presslight} optimizes rewards but cannot consider all relevant intersections as a whole, making global optimization difficult. PNC-HDQN~\cite{zhang2022neighborhood} separates the associated calculations between intersections from the training of RL agents, enabling RL algorithms to handle dynamic data changes. Zhang et al.~\cite{zhang2021independent} proposed a neighborhood cooperative Markov game framework. This framework assigns the objective of each intersection to be the average cumulative payoff of its neighborhood and employs a forgetful experience mechanism to reduce the importance of past experiences. Moreover, the framework can independently learn cooperative strategies based on the `generous' principle. However, fully considering the individualization of agents is infeasible in large-scale traffic scenarios, as it leads to inefficient model training. The leader-following paradigm~\cite{leader-following} is also an effective method for distinguishing between heterogeneous agents. Luan et al. used this model to distinguish roles in ATSC scenarios with two types of agents controlling intersections in traffic scenarios. However, the traffic flow distribution at each intersection is often different, and distinguishing roles may not account for the specific characteristics of each intersection.

In summary, traffic signal control (TSC) is essentially a multi-agent problem in which each intersection can be treated as an independent agent. In these complex systems, the coordination mechanism between agents is critical, which not only needs to consider the dynamic influence of neighbor intersections but also needs to accurately grasp the traffic flow characteristics of different periods within the same intersection. The transportation network in the real world often covers a large number of intersections, and these intersections cannot be simply treated as uniform nor considered in isolation. Therefore, we train a set of value functions by a single critic network with multi-head outputs shared by all intersections and dynamically adjust the weights of these value functions with the help of hyper-action output from the actor with a network to generate a value function with preference for each intersection. 

\section{Preliminary}
\label{sec:Preliminary}

\subsection{Partially Observable Markov Decision Process}

\begin{figure}[!t]
    \centering
    \includegraphics[width=12cm]{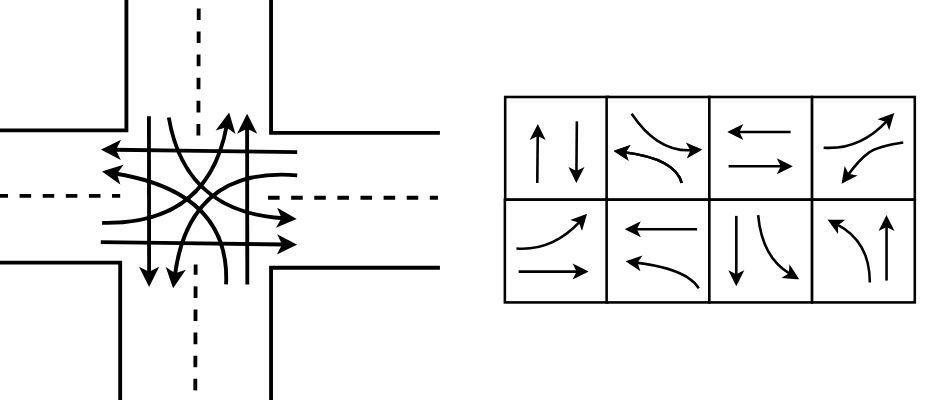}
    \caption{Signal phase and corresponding action set of crossroads.}
    \label{fig:action}
\end{figure}

In this section, the traffic signal control problem is described as a partially observable Markov decision process using the basic problem defined in multi-intersection traffic signal control~\cite{wei2019colight}. Each intersection in the system is controlled by an RL agent, which observes a portion of the entire system. The goal of each agent is to adaptively control the phase change of the signal lights to minimize the cumulative number of waiting vehicles at the intersection where it is located for a long time.  
Define the problem as a tuple: $<S,O,A,P,R,\gamma>$.
\begin{itemize}
    \item System state space $S$ and observation space $O$. Suppose there are $N$ intersections in the system, and $S$ is the state space used to describe all possible system states. 
    $\{O_1, . . . , O_{|N|}\}$ is the observation space, and $O_i$ of agent $i$ is observed partially from the state of the system. The observation space should reflect the characteristics of the road network as precisely as possible and in line with real-world constraints. In a TSC scenario, the number of vehicles waiting in the incoming lanes can be an indication of the traffic situation in time. Therefore, the observation is defined as: $o_i^t={wave[l]^t}_{l\in L_i}$. where $L_i$ is the set of incoming lanes at intersections $i$. ${wave[l]^t}$ is the number of waiting vehicles in the incoming lane $l$ at time $t$.
    
    \item Action space $A$. $\{A_1, . . . , A_{|N|}\}$ is the joint action space of all agents. At each time $t$, agent $i$ selects an action $a_i^t \in A_i$ as a decision for the next period. In this work, we follow the standard definition of action space, which is defined as a set of phases, \ie, a set of non-conflicting traffic movements. A traffic movement is a pair of one incoming lane and one outgoing lane. At the intersection shown in Figure \ref{fig:action}, there are eight phases in total.
    
    \item Transition probability $P$. The transition probability $P(s^{t+1}|s^t, \textbf{a}^t):S \times A_1\times\cdot\cdot\cdot\times A_N \times S \to [0,1]$ is the transition probability space that assigns a probability to each state-action-state transition. 
    
    \item Reward $R$. The reward $R_i$ of each agent $i$ is obtained by the reward function $R_i: S \times A_1\times\cdot\cdot\cdot\times A_N \to \mathbb{R} $. The optimization goal in ATSC is to shorten the travel time of vehicles, which is difficult to quantify directly with the reward of each intersection in each step. Therefore, we use the sum of the waiting queue lengths of vehicles in all lanes of the intersection as an alternative objective. In detail, the reward of agent $i$ at time $t$ is formalized by $r_i^t=-\sum_{l\in L_i} wave[l]^t$. 
    
    \item Policy $\pi$ and  discount factor $\gamma$. At time $t$, each agent chooses an action following a certain policy $\pi$. The aim is to learn an optimal joint policy $\pi^*= \pi_1^* \times \cdot\cdot\cdot\times \pi_N^*$ to maximize the cumulative discounted average reward $R^t=E[\frac{{1}}{{|N|}}\sum_{i\in N}\sum_{t=\tau}^T \gamma^{t-\tau} r_i^t]$, and $\gamma\in [0,1)$ distinguish rewards according to temporal proximity.
\end{itemize}

It should be noted here that in the model defined above, each agent has its own reward, which is different from the common multi-agent cooperation problem, where all agents share one reward. Obviously, the POMDP defined in this article is a mixed game problem, in which multiple agents with different individual rewards cooperate with each other to maximize their average return. There are two reasons why we define TSC issues in this way. Firstly, explicitly designing rewards for each agent helps improve the learning efficiency of the algorithm, as it does not need to face the credit assignment problem in cooperation MARL scenarios. In addition, we have also demonstrated the rationality of this design in our previous works~\cite{zhang2022neighborhood, zhang2021independent}. Previous works show that there is a weak cooperative relationship between multiple intersections in the same road network, and their individual returns are positively correlated. Therefore, optimizing individual returns will also optimize the overall average return.

This environmental setting directly leads to the personalized needs of different agents. Because different agents in the same system state have different rewards, the value function estimates for each agent under the same joint policy should also be different, which further leads to personalized requirements for individual policies. As described as above, we want all agents to share the same network to improve learning efficiency. The next question is how to design a MARL algorithm so that multiple agents with shared policy network can efficiently learn the optimal joint policy, while also ensuring the personalized needs of all agents.

\subsection{PPO}
PPO (Proximal Policy Optimization)~\cite{schulman2017proximal} is a popular reinforcement learning algorithm that combines the advantages of policy optimization and value estimation methods. It aims to effectively optimize policies while ensuring stability and convergence by limiting the difference between the new policy and the old policy in each update, thereby preventing the policy from deviating too far during training. Specifically, PPO introduces a clipping mechanism to limit the updating steps of policy and optimizes the policy parameter $\theta$ by maximizing the PPO-clip objective of 
\begin{equation}
\label{eq:a_loss}
\begin{split}
{L^{{\rm{CLIP}}}}(\theta ) = E\left[ {min({\rho _t}(\theta ){A_t},clip({\rho _t}(\theta ),1-\varepsilon ,1 +\varepsilon){A_t})} \right]
\end{split}
\end{equation}
where $\theta$ is the policy parameter, $E$ denotes the empirical expectation over timesteps. $\rho _t$ is the importance sampling ratio, \ie, the ratio of the probability under the new and old policies formalized by $\rho _t = \frac{{{\pi _\theta }({a_t}|{s_t})}}{{{\pi _{{\theta _{old}}}}({a_t}|{s_t})}}$. $A_t$ is the estimated advantage at time $t$ and $\varepsilon$ is a hyperparameter. The function $clip(a,b,c)$ clips the value $a$ to the interval $[a, b]$. In this article, we propose a centralized training and decentralized executing (CTDE) MARL algorithm based on the PPO framework, where all agents share parameters but have different preferences.

\section{Hyper-Action Multi-Head Proximal Policy Optimization}
\label{sec:HAMH-PPO}

\begin{figure}[!h]
    \centering
    \includegraphics[width=14cm]{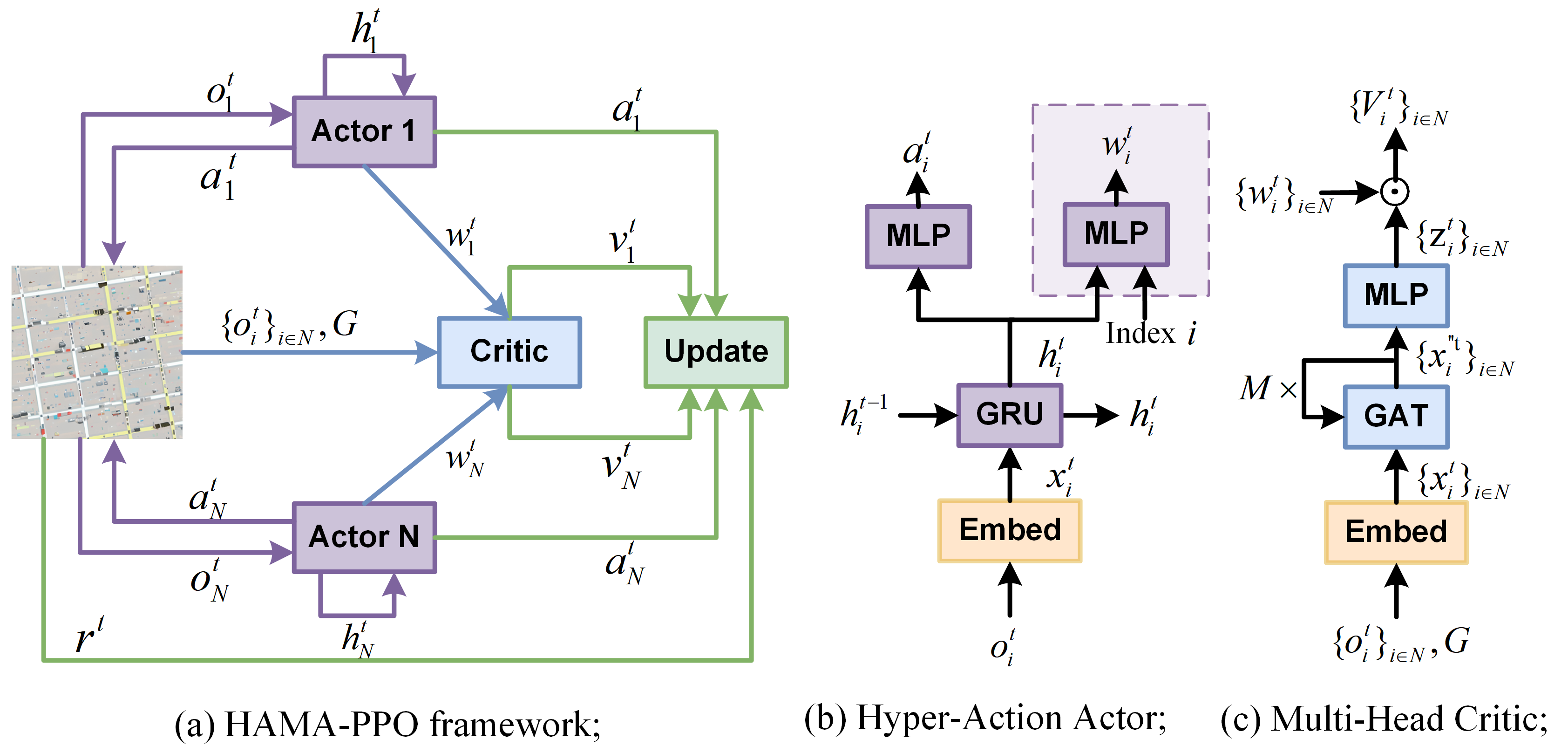}
    \caption{The overall structure of HAMH-PPO is illustrated in Figure (a) and is designed based on the PPO algorithm. It consists of N agents, each corresponding to an intersection, and a centralized critic. Figure (b) shows the structure of a single actor, where local observations are passed through feature extraction and output action $a_i^t$ that affects the environment, while cyclic features combined with the intersection index $i$ output the hyper-action $w_i^t$. Figure (c) represents the structure of the critic, where global observations and graph information are processed using GAT to generate a set of value functions for each intersection. The joint value function, obtained by dot product with the hyper-action, estimates the average travel time at the current intersection.}
    \label{fig:Arch}
\end{figure}

In this section, we provide the implementation details of the Hyper-Action Multi-Head Proximal Policy Optimization (HAMH-PPO) algorithm. Fig.\ref{fig:Arch}(a) shows the overall framework of HAMH-PPO. HAMH-PPO consists of a shared decentralized actor Fig.\ref{fig:Arch}(b) and a shared centrally trained critic Fig.\ref{fig:Arch}(c), and employs the CTDE paradigm to train agents to learn coordinated policies. Unlike the classical cooperative CTDE algorithm MAPPO, our algorithm provides estimates of the value function for each intersection, and the centralized training mainly involves training with global data combined together, for the reason that we need to estimate different state value for agents with different individual returns.

Specifically, in a TSC scenario with $|N|$ shared network parameter (a shared actor network and a shared critic network) agents, where all agents make decisions based on their local observation history by actor network and estimate their state values by a centralized critic network. The centralized critic network, shown in Fig.\ref{fig:Arch}(c), is shared by all agents, estimating returns for each agent by accessing global information. Considering the obvious characteristics of the graph structure between intersections in TSC scenarios, the critic uses a graph attention neural network to calculate the graph representations of all intersections (nodes in the graph) in the road network. This graph representation is further processed through a shared MLP layer to output the state value of each intersection (agent). In addition, as shown in Fig.\ref{fig:1_3_Observed_distribution}, it is noted that there are differences in the observation distribution and state transitions of different intersections in the road network, and the sampling frequency of the corresponding data for the observation representation of different agents is different, resulting in different returns. Even if graph neural networks can output different observation representations for different agents from one global system state, it is insufficient to use a single linear mapping (\ie, MLP layer) to estimate the values of all agents. To this, the critic network in HAMH-PPO outputs a set of value functions $z_i^t$ with set size $|z_i^t|=k$ for each agent which is implemented by an MLP layer with multiple heads. The decentralized execution actor network, shown in Fig.\ref{fig:Arch}(b), is also shared by all agents. They take the local observation history as input and extract the time characteristics of the intersection through GRU. The outputs are distributions of action $a_i^t$ and hyper-action $w_i^t$ of agent $i$. The action distribution is the same as the standard RL definition, which outputs the probability of selecting a certain action $a_i^t$ in the action space, and the hyper-action $w_i^t$ is also a probability used to balance the multiple values estimated from the centralized critic using a dot-product operation and further guide the updating of action policies. So we call the actor and the critic networks hyper-action actor (HA-Actor) and multi-head critic (MH-Critic) respectively. The following are the specific details of the two networks.

\subsection{Hyper-action actor network (HA-Actor)}
In the traffic signal control problem, historical observations are critical. Therefore, HA-Actor uses the GRU to handle long sequential data to capture the temporal dependency at intersections. As can be seen from Fig.\ref{fig:Arch}(b), for each agent $i$, the actor network's input is the observation $o_i^t$ and historical representation $h_i^{t-1}$. A multi-layer perception is used to obtain the observation embedding $x_i^t$, followed by a GRU unit to capture a temporal representation of the current state of agent $i$, formally,
\begin{equation}
   x_i^t = Embed(o_i^t)=ReLU \lbrack W_c\cdot ReLU(W_d\cdot o_i^t +b_d) + b_c) \rbrack  
\end{equation}%
\begin{equation}
   h_i^t=GRU(x_i^t, h_i^{t-1})     
\end{equation}%
where $W_c, W_d, b_c$ and $b_d$ are the weight matrixs and bias vectors to learn. The hidden feature $h_i^t$ outputs the action probability distribution through the linear layer on the left, and then selects an action $a_i^t$ based on the probability sampling.

In HAMH-PPO, the core idea of personalization is personalized learning through generating distinct value estimation guidance policies for intersections. Compared to a single value function, a set of value functions can select different estimates for different intersections. However, selecting an estimation value function from a finite set of value functions for the current action may not be sufficient. One idea is to consider whether we can permit several value functions to jointly guide the update of the policy network. A hyper-action network is hence introduced, which takes the hidden features $h_i^t$ and intersection index $i$ as input to generate weights $w_i^t$ for each value function. We can provide numerous combinations of value functions, thereby liberating the solution space from constraints.
It is noteworthy that $w_i^t$ is a probability distribution of the value functions, rather than a multidimensional action. For the sake of convenience, we will designate the dashed portion in Fig.\ref{fig:Arch} as $hp_\varphi$. 
\begin{equation}
   w_i^t(\cdot|o_i^t,h_i^{t - 1},i) = softmax((h_i^t \oplus i){W_e} + {b_e})    
\end{equation}%
where $W_e$ and $b_e$ are the weight matrix and bias vector to learn, $h_i^t$ is the historical representation at time $t$ output from the GRU unit. $\oplus$ is the concat operator.

The algorithm adopts a centralized training method to train actors. Due to parameter sharing, we can integrate the local observation data of all agents to jointly train the actor network. Given a trajectory of interactive data ${\left\{ {\left\langle {o_i^t,a_i^t,o_i^{t + 1},r_i^t} \right\rangle } \right\}_{i \in N}^{t<T}}$, the hyper-actor trains its network parameters $\theta$ based on the following averaged PPO~\cite{schulman2017proximal} loss, 
\begin{equation}
\label{eq:a_loss}
\begin{split}
{L^{PPO}(\theta ) = \frac{1}{T|N|}\sum\limits_{t < T} {\sum\limits_{i \in N} {min(\rho _i^t(\theta )A_i^t,clip(\rho _i^t(\theta ),1 - \varepsilon ,1 + \varepsilon )A_i^t)} } }
\end{split}
\end{equation}
where ${A_i^t}$ is the advantage value calculated from the Generalized Advantage Estimation (GAE)~\cite{DBLP:journals/corr/SchulmanMLJA15}. $\rho_i^t$ is the importance sampling ratio. In addition, to prevent the slow update of certain value functions due to excessively small individual weights, we added an entropy regularization term for the hyper-action distribution, \ie, $H(w_i^t(\theta))$, where $w_i^t(\theta)$ denotes the $w_i^t(\cdot|o_i^t,h_i^{t - 1},i)$ parameterized by $\theta$. In conclusion, the current loss function of the actor network is:
\begin{equation}
\label{eq:actor_loss}
    L(\theta)=L^{PPO}(\theta ) -\lambda H(w_i^t(\theta))
\end{equation}
where $\lambda$ is the weight entropy coefficient. The use of entropy regularization term in the new loss can enable the distribution of hyper-action to learn in the direction of increasing entropy, preventing it from tending towards a deterministic distribution prematurely.

\subsection{Multi-head critic network (MH-Critic)}

\begin{figure}[h]
    \centering
    \includegraphics[width=12cm]{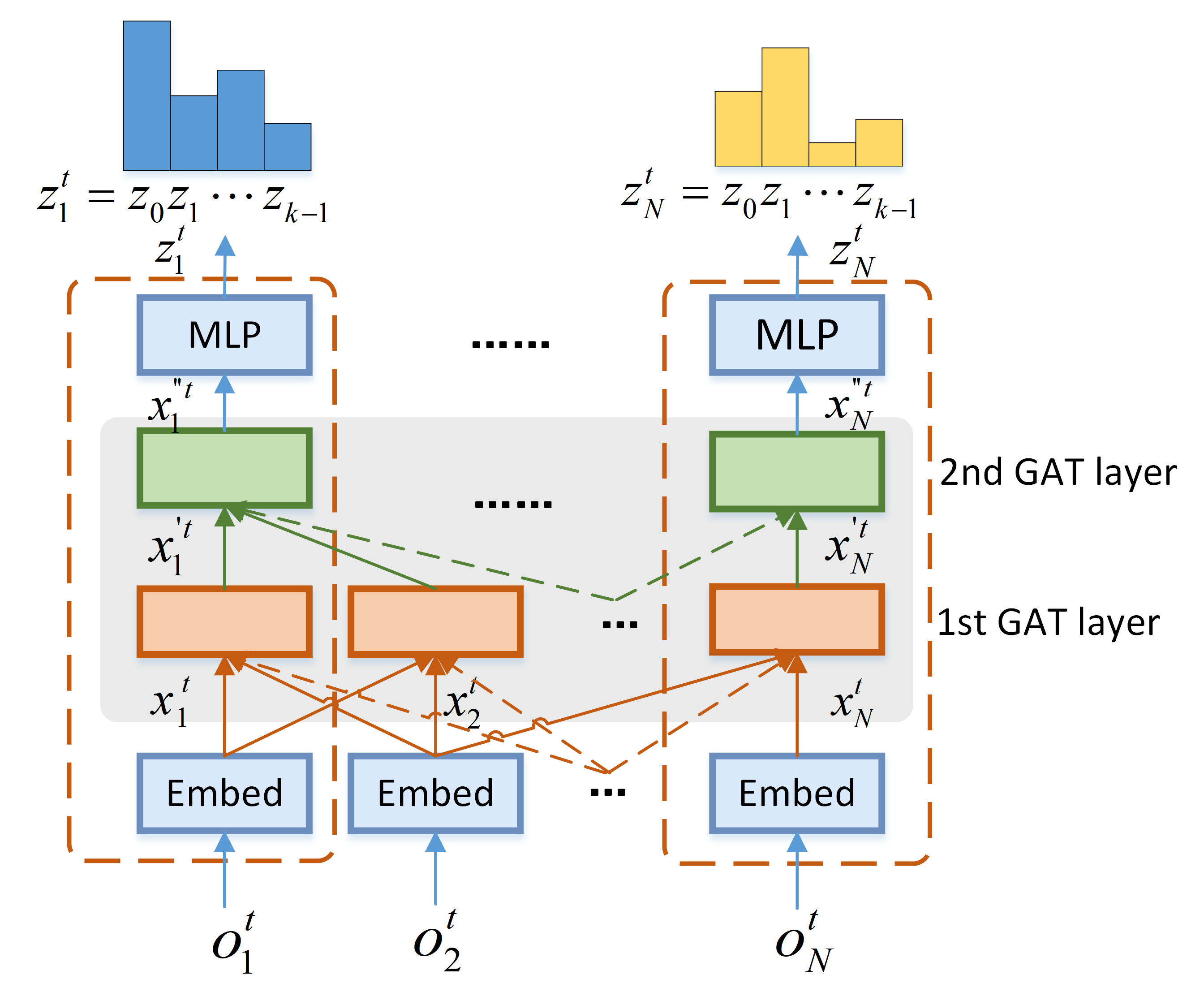}
    \caption{The detailed structure of the cubic network, which ultimately outputs $k$-dimensional value estimates.}
    \label{fig:GAT}
\end{figure}

The centralized critic needs to access global information, which is composed of the local observations of all agents and the adjacency relationships between graph nodes. We employ a Graph Attention Network (GAT) to process the structural information of the graph and extract spatial features between intersections. Fig.~\ref{fig:GAT} illustrates the detailed structure of the critic network, which includes $N$ agents. The local observations of each agent are processed through a multi-layer perception to obtain the observation embedding. To improve the overall system's generalization ability and performance, the actor and critic share an observation embedding layer, which not only reduces the number of parameters in the network but also promotes better information sharing and complementarity between them. The observations of all agents are fed through an embedding layer to obtain the embedding features $\{x_i^t\}_{i\in N}$ of each intersection. 

As shown in Fig.~\ref{fig:GAT}, the embedding features pass through a layer of GAT to combine neighbor information based on their importance and obtain ${x}_i^{'t}$. To increase the perceptual range of agents, we use multiple layers of GAT. In addition, more attention heads focus on different feature subspaces to acquire more relational representations, making the training more stable. The extracted spatial features undergo dimensionality reduction and aggregation through multiple perceptions, ultimately resulting in a vector $z_i^t$. Each intersection has its value function vector, representing multiple reward estimates for the current state. The set of value functions are used as experts to guide the training of agents' policies, and our core idea is to make these experts learn in parallel and cooperatively to guide the updates of the policy network. Here, the hyper-actions $w_i^t=w_0...w_{k-1}$ represent the probability distribution of data items in $z_i^t$. Our experiments show that as the differences in traffic conditions at intersections increase, the disparities among these weights also grow. As a result, the proportions of value functions in the policy updates vary, and thus personalized learning of the policy can be achieved. Multi-valued functions and hyper-action have the same dimension $k$, and their dot product produces the final joint value function, providing precise and comprehensive guidance for optimizing the policy network. Formally, the value function of agent $i$ at time $t$ can be calculated by following,
\begin{equation}
    V_i^t = z_i^t(\cdot|{\{ o_i^t\} _{i \in N}},G) \odot w_i^t(\cdot|o_i^t,h_i^{t - 1},i)
\end{equation}
where $\odot$ is the dot product operator of two vectors. $z_i^t$ and $w_i^t$ are value distribution and hyper-action of agent $i$ respectively. $G$ is the connectivity matrix between intersections used for GAT, which is fixed as a constant matrix in each traffic scenario in this article.

The critic loss function is consistent with PPO~\cite{schulman2017proximal}, \ie, the TD squared error. Using a network $\varphi$ to parameterize $V$, then the loss function of the MH-Critic can be defined by following
\begin{equation}
\label{eq:critic_loss}
    L(\varphi)=\frac{1}{2NT}\sum_{t<T}\sum_{i\in N}[r_i^t+\gamma V_i^{t+1}(\varphi) -V_i^{t}(\varphi)]^2.
\end{equation}

Algorithm 1 shows details of HAMH-PPO. HAMH-PPO trains two neural networks that all agents share: an actor network $\theta$ and a critic-network $\varphi$. For each episode, the agents interact with the environment by executing their joint policy and collecting a set of trajectories (Step 4). The trajectories are stored in a replay memory $D$ (Step5), and all networks are trained based on Eq.\ref{eq:actor_loss} (Step 8) and Eq.\ref{eq:critic_loss} (Step 9). To increase data utilization and improve sample efficiency, we train $E$ epochs on the same trajectory.

\begin{algorithm}[tb]
    \caption{HAMH-PPO}
    \label{alg:algorithm}
    \textbf{Input}: Episodes $K$, Number of agents $N$, Time steps per episode $T$, Epoch $E$.\\
    \textbf{Output}: Trained critic $\varphi$ and actor $\theta$.
    \begin{algorithmic}[1] %[1] enables line numbers
        \STATE Initialize critic-network $\varphi$, actor network $\theta$, Replay buffer $D$.
        \FOR {$k=0,1,...,K-1$}
        \STATE Reset simulator environment.
        \STATE Set data buffer $D=\phi$
        \STATE Collect a set of trajectories by running the policy $\pi_\theta$.
        \STATE Store ${(o_i^t,a_i^t,r_i^t,o_i^{t+1})}_{i\in N, t\in \{0,...,T-1\}}$ in buffer $D$
        \FOR{$e=0,1,...,E-1$}
        \STATE For each trajectory from $D$.
        \STATE Update $\theta$ by Eq.\ref{eq:actor_loss}.
        \STATE Update $\varphi$ by Eq.\ref{eq:critic_loss}.
        \ENDFOR
        \ENDFOR
    \end{algorithmic}
\end{algorithm}

\section{Experiments}
\label{sec:Experiments}

In this section, we conduct extensive experiments to answer the following questions:
\begin{itemize}
    \item RQ1: How does our proposed method perform compared to other advanced methods?
    \item RQ2: Can HAMH-PPO  generate personalized value estimates in road networks?
    \item RQ3: How do the components used in HAMH-PPO affect the performance of the algorithm?
\end{itemize}
\subsection{Settings}

We conducted experiments on the CityFlow traffic simulator. CityFlow is an open-source simulator widely used for traffic signal control at multiple intersections. After inputting traffic data into the simulator, the vehicle moves toward its destination according to environmental settings. The simulator provides status to the signal control method and executes traffic signal actions from the control method. Typically, there is a three seconds yellow light signal and a two-second full red light signal after each green light signal.

\begin{table}[!t]
    \centering
    \caption{Scenarios statistics.}
    \begin{tabular}{c|ccccc}
        \hline
        \multirow{2}*{Scenarios}& \multirow{2}*{Intersections}& \multicolumn{4}{c}{Arrival num(vehicles/300s)}\\
         \multirow{2}*{}&\multirow{2}*{}& Mean & Std & Max &	Min \\
        \hline
        $Grid_{10\times10}$&100&880&0&880&880\\
        \hline
         $Grid_{4\times4}$&16&935.92&17.47&960&896 \\
         \hline
         $D_{Jinan1}$& 12&645.42&3.45&652&639\\
         \hline
         $D_{Jinan2}$& 12&645.75&4.75&654&639\\
         \hline$D_{Jinan3}$& 12&419.67&98.53&672&256\\
         \hline
         $D_{NewYork}$&48&235.33&5.59&244&224\\
         \hline
         % \toprule
    \end{tabular}
    \label{tab:dataset_road}
\end{table}

\subsection{Dataset}
We conducted algorithm validation in a total of 6 scenarios (Jinan $\times $ 3, New York $\times $ 1, synthetic road networks $\times $ 2) on two real road networks and two synthetic road networks. Table~\ref{tab:dataset_road} shows the distribution of traffic flow in each scenario. In all scenarios, each car has its own parameters, such as acceleration and maximum speed. Each road has three lanes, one left turn lane and two through lanes. We set the phase number as eight and the minimum action duration as 10 seconds. 

\textbf{Synthetic Data.} In the synthetic dataset, we will use two kinds of maps. They are made up of different numbers of intersections. Each road at the intersection has three lanes with 3 meters in width. Synthetic maps are generated via Cityflow and the traffic of the two road networks has a period of 3600 seconds.
\begin{itemize}
    \item $Grid_{10\times10}.$ The road network has 100 ($10\times10$) intersections, each of which is four-way. The road network structure is irregular. The east$\rightarrow$west traffic flow is 900 vehicles/lane/hour and 150 vehicles/lane/hour, and the north$\rightarrow$south traffic flow is 720 vehicles/lane/hour and 90 vehicles/lane/hour.
    \item $Grid_{4\times4}.$ The road network has 16 ($4\times4$) intersections. Each intersection is four-way, with two 300-meter(East-West) road segments and two 300-meter (South-North) road segments. The traffic flow is not regular. 
\end{itemize}

\textbf{Real-world Data.} We also use real-world traffic data from two cities: Jinan and New York. Their road network structure is imported from OpenStreetMap. The traffic of the two road networks has a period of 3600 seconds. Jinan includes three traffic flow distributions with different traffic arrival rates. 
\begin{itemize}
    \item $D_{Jinan}.$ The road network has 12 ($3\times4$) intersections. Each intersection is four-way, with two 400-meter(East-West) road segments and two 800-meter (South-North)road segments. In our experiment, this road network has three datasets settings. 
    \item $D_{Newyork}.$ The road network has 48 ($3\times16$) intersections. Each intersection is four-way, with two 350-meter(East-West) road segments and two 100-meter (South-North)road segments.
\end{itemize}

\subsection{Baseline}
Our experiment mainly compares two types of methods, traditional traffic signal control methods and signal control methods based on RL. For a fair comparison, all RL methods select phase from the set of actions. Table~\ref{tab:parameters} shows parameter settings used in each RL algorithm. The details are as follows:
\begin{itemize}
    \item \textbf{Fixedtime\cite{FixedTime_2008}.} Traffic signals at intersections operate according to a predetermined timing scheme, with traffic signals changing periodically.
    \item \textbf{MaxPressure\cite{Varaiya_2013}.} The objective is to minimize the phase ``pressure" at intersections by balancing the queue lengths between adjacent intersections, to minimize the phase pressure at each intersection to maximize the throughput of the entire road network.
    \item \textbf{PNC-HDQN\cite{zhang2022neighborhood}.} A fully decentralized neighborhood learning framework, where each agent preprocesses neighborhood data based on the correlation between two intersections.   
    \item \textbf{MA2C\cite{chu2019multi}.} A fully decentralized and scalable MARL algorithm where each local agent is optimized through coordination between multiple intersections.
    \item \textbf{MPLight\cite{chen2020toward}.} A large-scale traffic road network algorithm using FRAP~\cite{DBLP:conf/cikm/ZhengXZFWZLXL19} as the base model to train parameter-sharing agents using DQN.
    \item \textbf{Colight\cite{wei2019colight}.} By introducing the graph attention network, the dynamic influence of surrounding intersections on the current intersection is considered.
\end{itemize}

\begin{table}[!h]
    \centering
    \caption{Main hyperparameters of HAMH-PPO.}
    \begin{tabular}{ll}
    \hline
         Parameter&  Value\\
         \hline
         Hidden state dimension of GRU & 128 \\
         Hidden state dimension of GAT & 128 \\
         Actor learning rate $\alpha_{\theta}$& 5e-4 \\
         Critic learning rate $\alpha_{\varphi}$& 5e-4 \\
         PPO clip $\varepsilon$ & 0.2\\
         PPO epoch $E$ & 15\\
         Discount factor $\gamma$ & 0.98\\
         Network optimizer & Adam\\
         Entropy coefficient $\lambda$ &0.01\\
         Dimension $k$ of hyper-action  & 32\\
         \hline
    \end{tabular}
    \label{tab:parameters}
\end{table}

\subsection{Evaluation Metric}

The goal of traffic signal control is to move vehicles through intersections more quickly, and the average travel time is the average time spent by all vehicles entering and exiting the road network. Therefore, we use average travel time to evaluate the performance of signal control algorithms. This is also the most commonly used performance metric in the transportation domain. The average travel time of all vehicles:
\begin{equation*}
    m_{tt} = \frac{1}{|v_{in}|}\sum_{v\in v_{in}}(t_v^{out}-t_v^{in}),
\end{equation*}
where $t_v^{out}$ and $t_v^{in}$ represent the entry and exit times of vehicle $v$, respectively, while $v_{in}$ denotes vehicles entering the road network.

\subsection{Performance Comparison (RQ1)}
In this section, we present the performance of HAMH-PPO, comparing it with the traditional transportation methods and RL methods in six scenarios. Figure \ref{tab:parameters} summarizes the hyperparameter settings in HAMH

\begin{table*}[]
    \centering
    \caption{Performance on synthetic data and real-world data.}
    \setlength{\tabcolsep}{1.1mm}{
    \begin{tabular}{c|cccccc}
        \hline
         Methods&$Grid_{4\times4}$ &$Grid_{10\times10}$& $D_{Jinan1}$ & $D_{Jinan2}$& $D_{Jinan3}$& $D_{Newyork}$  \\
         \hline
         FixedTime  &725.11 &1440.88 &537.64 &537.76 &514.14	 &1122.95\\
         MaxPressure&431.76 &953.98 &798.89   &694.90&374.56  &203.08\\
         \hline
         PNC-HDQN   &512.64 &1170.48 &727.73 &584.65&374.78   &1187.89 \\
         MA2C       &677.12 &1199.02 &970.6&825.25&445.91    &1215.44 \\
         MPLight    &1047.59&1405.92 &480.02 &430.84&348.21   &194.23 \\
         Colight    &389.02 &1100.49 &635.55 &502.32&335.89   &186.49 \\
         \hline
         \textbf{HAMH-PPO}&\textbf{273.90}&\textbf{669.34}&\textbf{427.44}&\textbf{366.26}&\textbf{306.45}&\textbf{181.69 }\\
         \textbf{HAMH-PPO}&\textbf{273.90}&\textbf{669.34}&\textbf{427.44}&\textbf{366.26}&\textbf{306.45}&\textbf{181.69 }\\
         \hline
    \end{tabular}
    \label{tab:performance}
    }
\end{table*}

\begin{figure}[!h]
    \centering
    \subfigure[$D_{Jinan1}$]{
    \label{fig:3_4}
    \includegraphics[width=6.5cm]{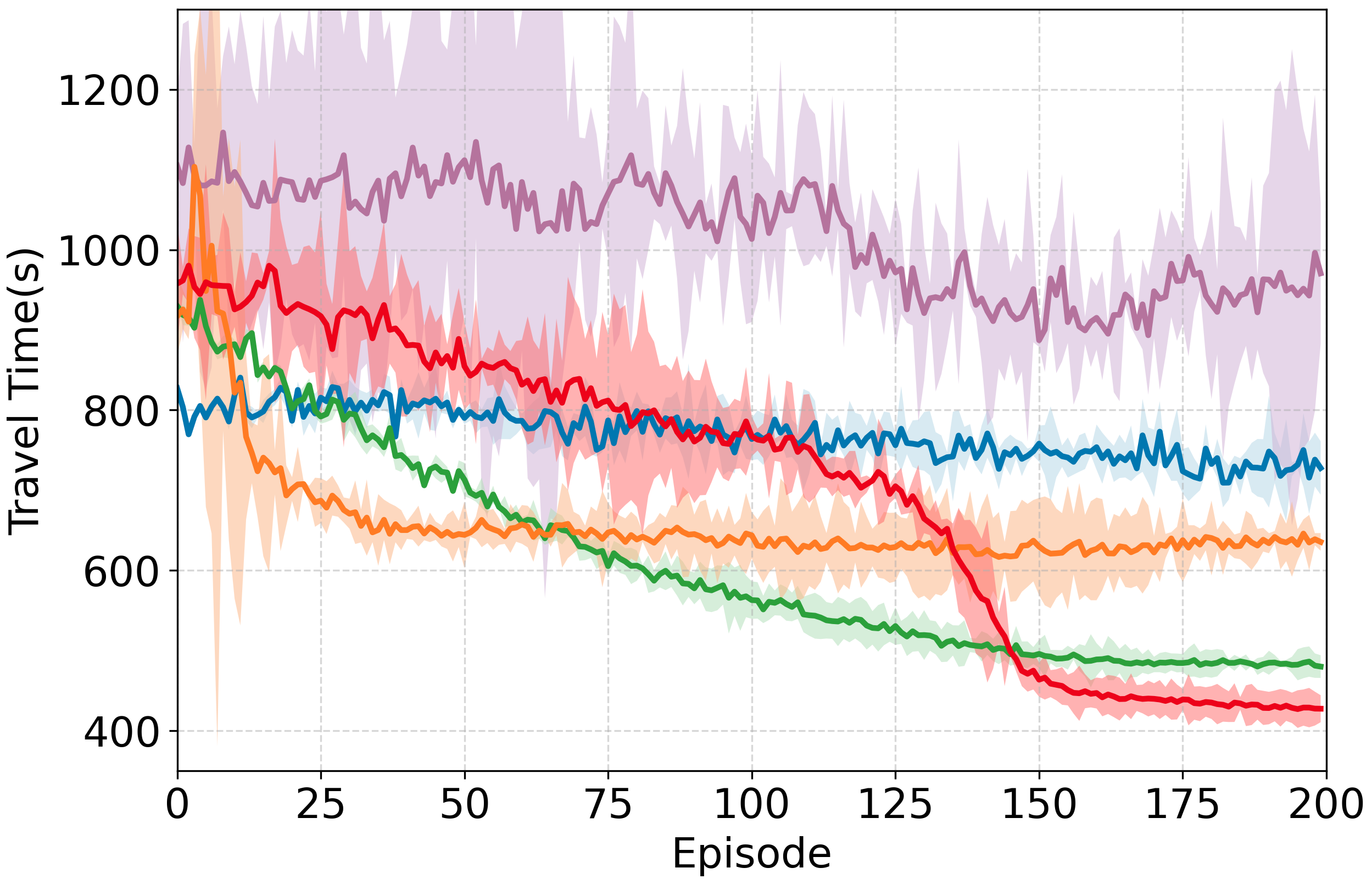}}
    \subfigure[$D_{Jinan2}$]{
    \label{fig:3_4}
    \includegraphics[width=6.5cm]{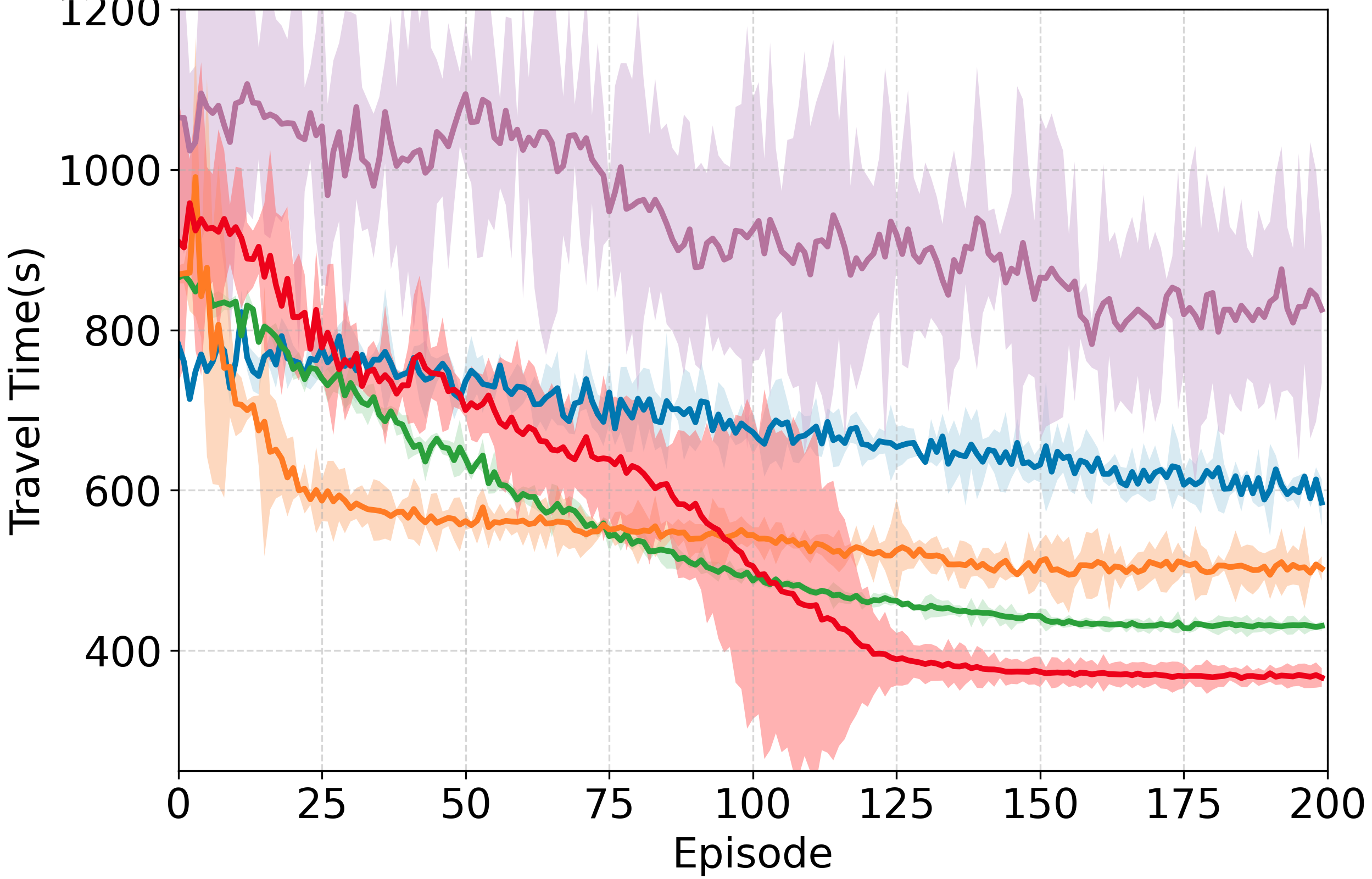}}
    \subfigure[$D_{Jinan3}$]{
    \label{fig:3_4}
    \includegraphics[width=6.5cm]{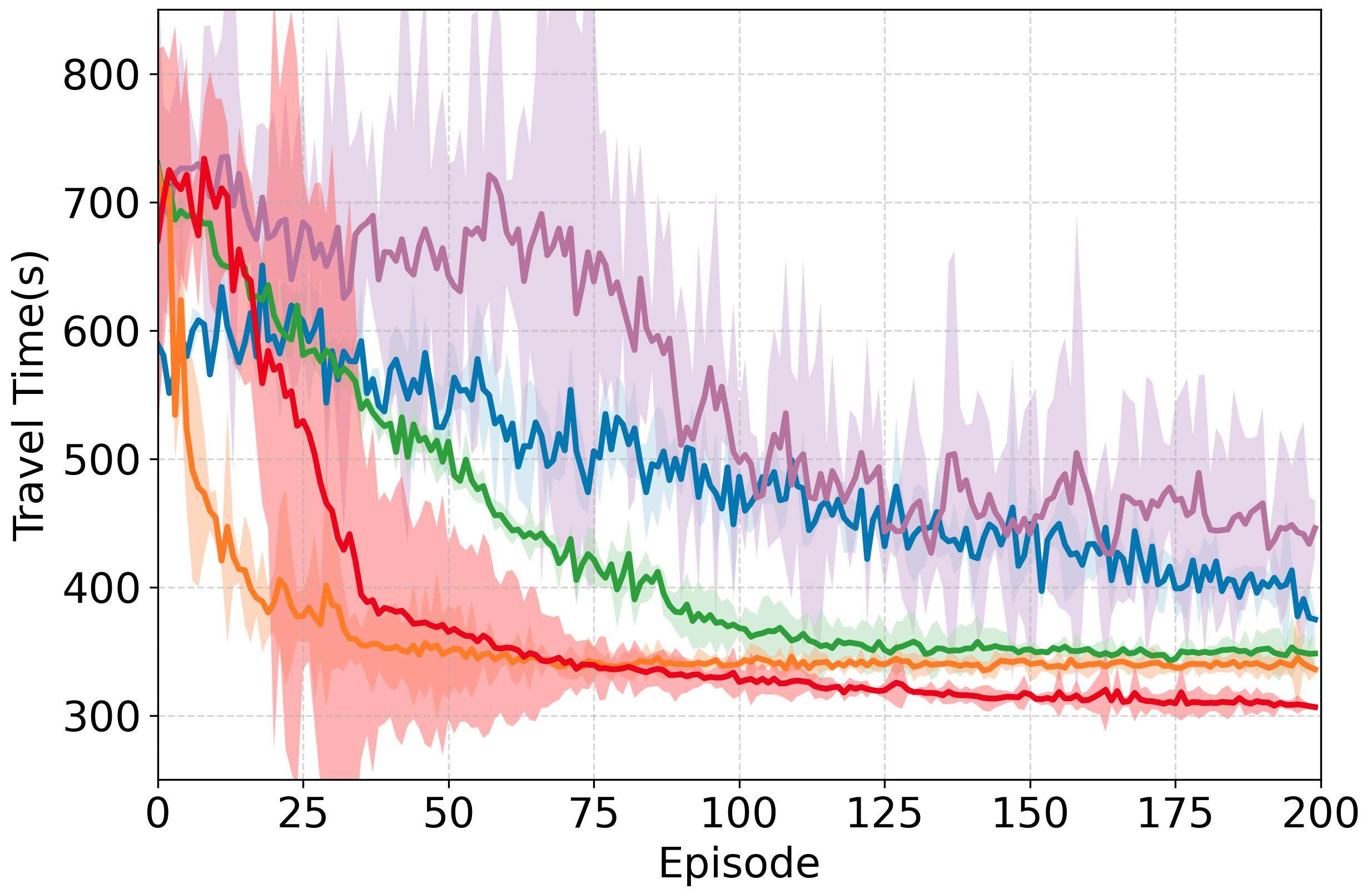}}
    \subfigure[$Newyork$]{
    \label{fig:3_16}
    \includegraphics[width=6.5cm]{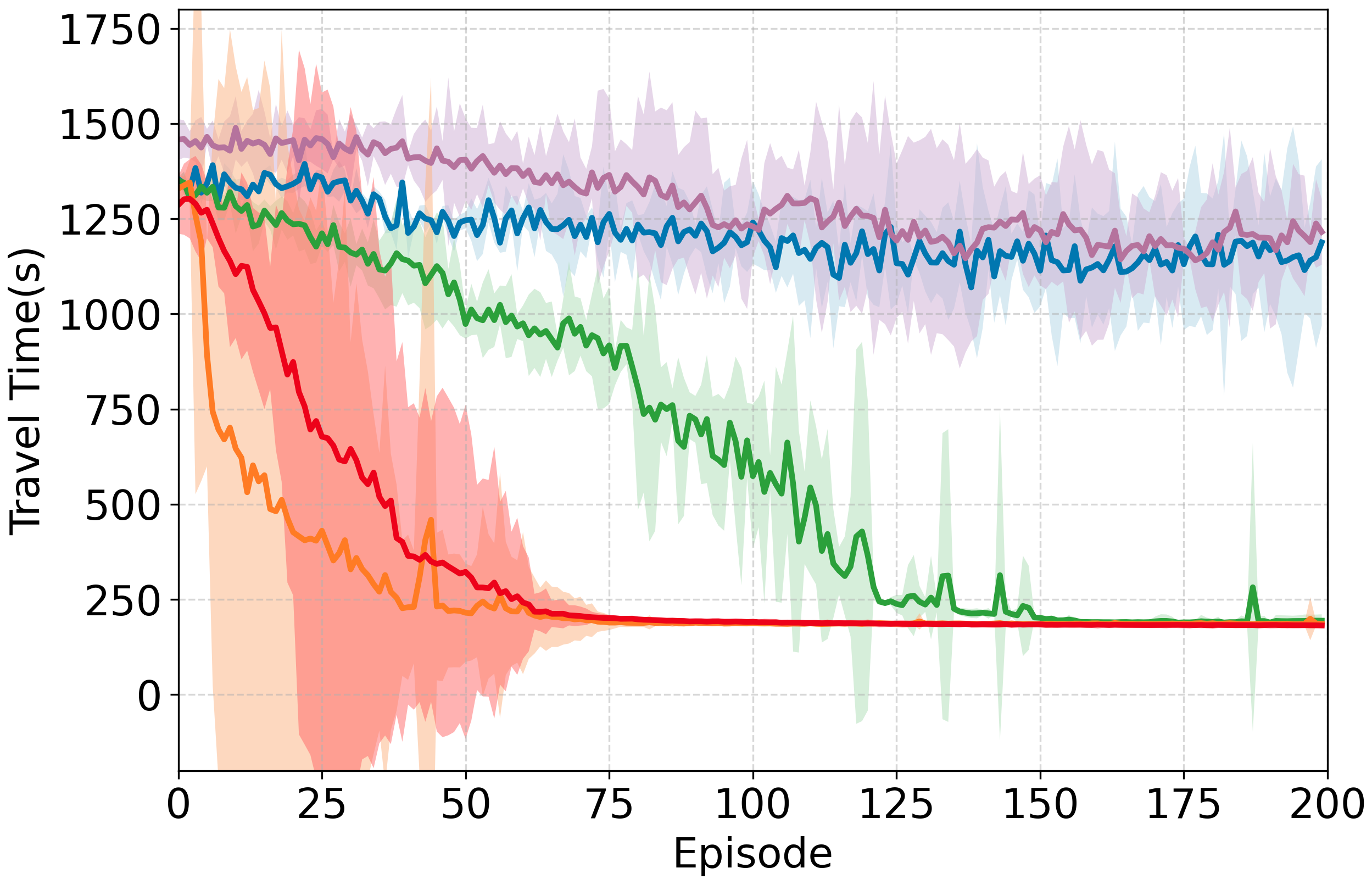}}
    \subfigure[$Grid_{4\times4}$]{
    \label{fig:4_4}
    \includegraphics[width=6.5cm]{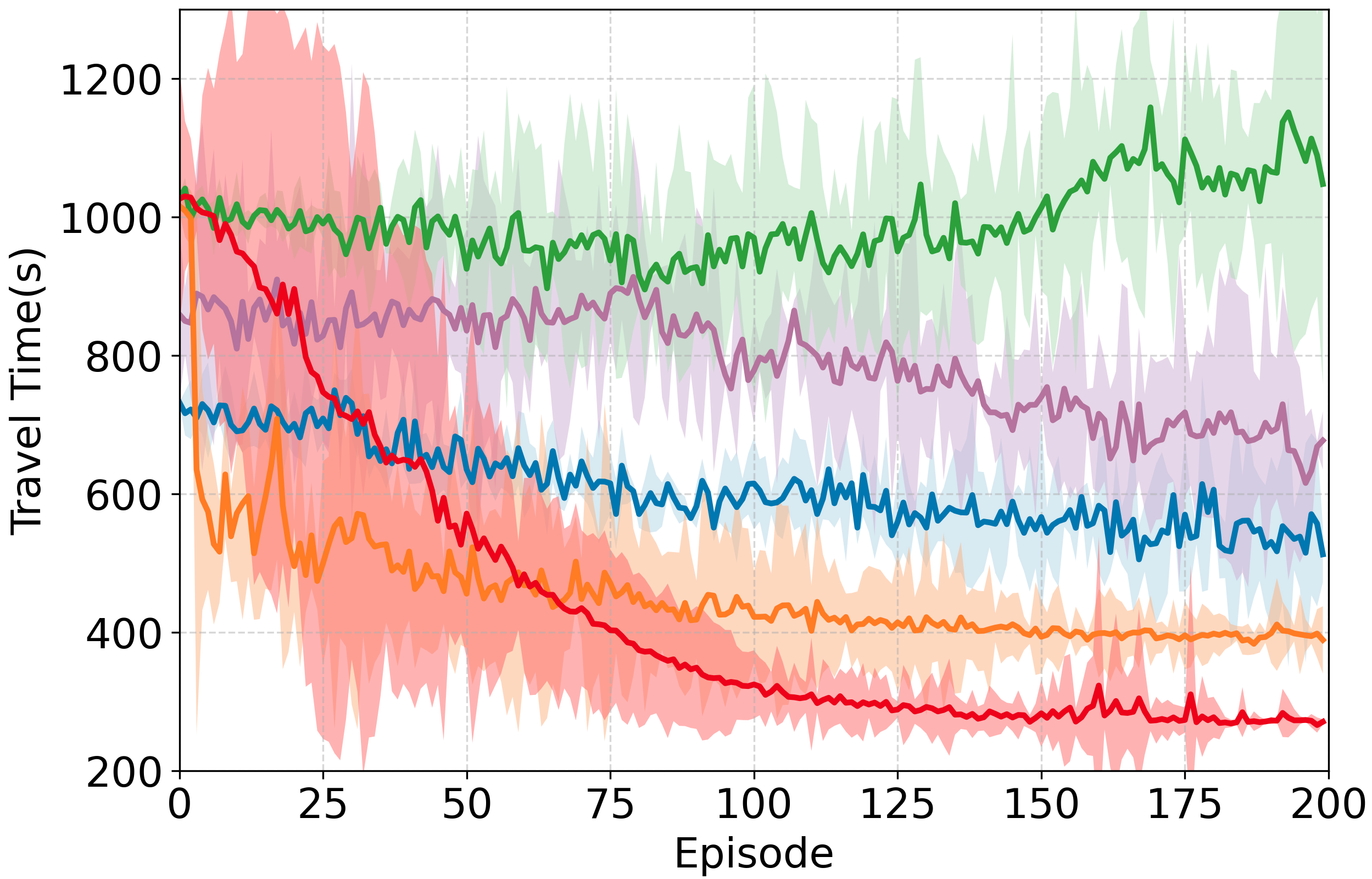}}
    \subfigure[$Grid_{10\times10}$]{
    \label{fig:10_10}
    \includegraphics[width=6.5cm]{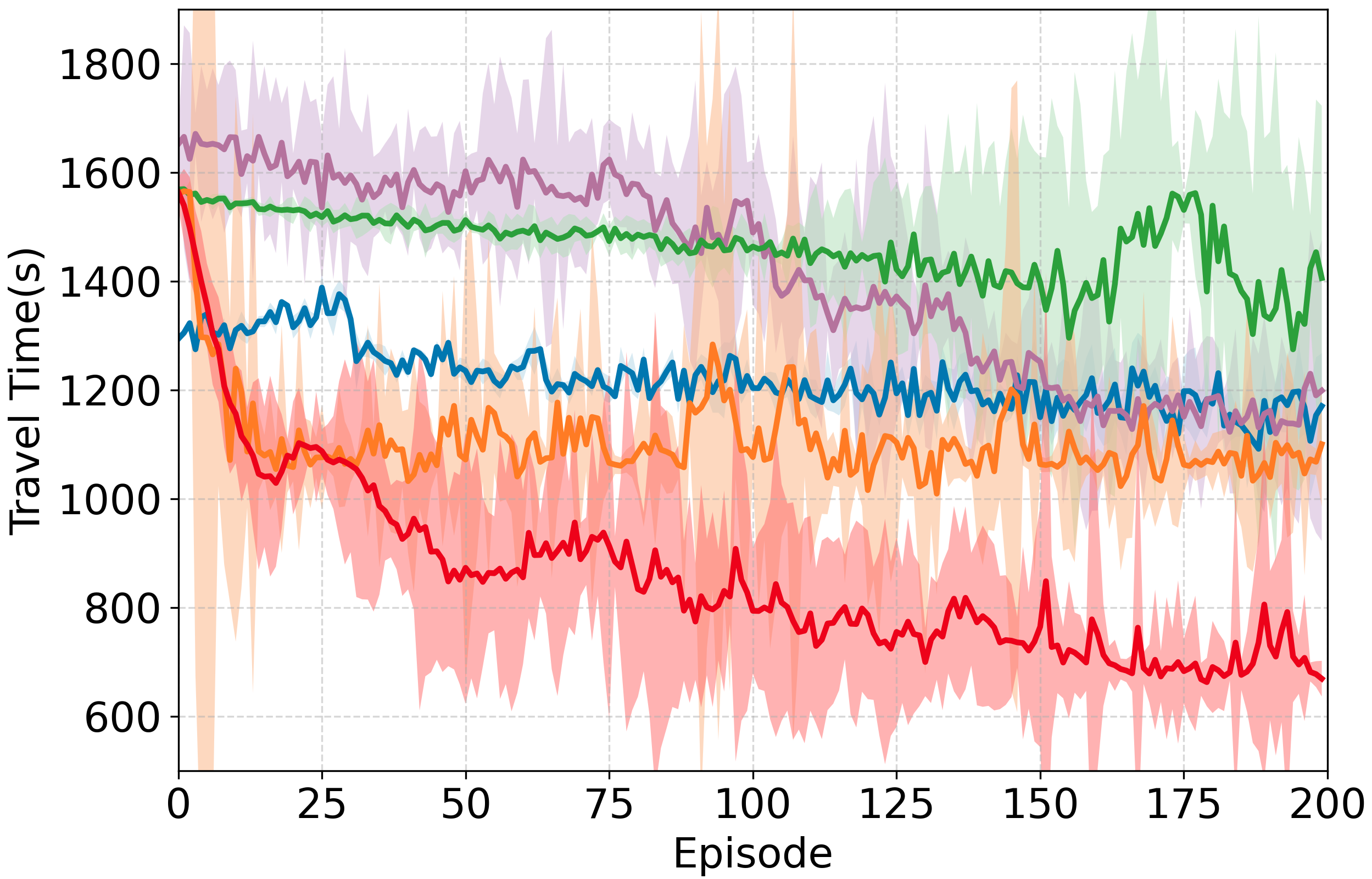}}
    \caption{Performance comparison of RL methods (PNC-HDQN `\textbf{{\color[HTML]{0077B0}---}}', MA2C `\textbf{{\color[HTML]{B5739D}{---}}}', Colight `\textbf{{\color[HTML]{FF7B24}{---}}}', MPLight `\textbf{{\color[HTML]{2BA03B}{---}}}' and HAMH-PPO `\textbf{{\color[HTML]{ED021A}{---}}}') in six datasets.}
    \label{fig:performance}
\end{figure}

Table \ref{tab:performance} compares HAMH-PPO and six other methods, including two traditional TSC methods and four advanced TSC methods for MARL. Compared to the two traditional methods, HAMH-PPO showed an average improvement of 48.67\% over Fixedtime and 31.42\% over MaxPressure across six datasets. This is because traditional traffic signal control is not suitable for traffic conditions that change over time. 

We also compare HAMH-PPO with four state-of-the-art MARL-based TSC methods. As can be seen from the table, our algorithm achieves the best results in all scenarios. Firstly, the traffic flow in synthetic road network 4$\times$4 is very large, and MPLight cannot deal with this situation without neighbor information. HAMH-PPO is highly effective, which is 29.6\% higher than Colight. Secondly, it can be seen that PNC-HDQN and MA2C are ineffective in the NewYork road network. These two algorithms train one agent for each intersection, which cannot deal with such large-scale complex real road networks. The final training results of MPLight, Colight, and HAMH-PPO are similar because there is less traffic distribution in the road network. In addition, since the synthetic 10$\times$10 road network is large and irregular, HAMH-PPO shows the advantage more prominently on this map than other maps. HAMH-PPO achieves a 29.8\% reduction in maximum travel time compared to the second-best scheme, which highlights the effectiveness of our method in large-scale TSC. The main reason is that our method trains personalized strategies and selects appropriate value estimates for different intersections in the training process, which enhances the representation ability of the trained model. The experimental results show that the model can significantly improve the efficiency of vehicle traffic. In summary, HAMH-PPO can effectively adjust the hyper-action according to the current road conditions, and generate a unique value estimate for each intersection. It achieves excellent performance in different road networks, especially in large-scale TSC.

In Fig.~\ref{fig:performance}, we compare PNC-HDQN, MA2C, MPLight, Colight, and HAMH-PPO’s convergence rates during training. The lines and shadows around the curves represent these algorithms' average travel time and error range on learning episodes. It is worth noting that MA2C and PNC-HDQN train an agent for each intersection, and the convergence behavior of the two methods is quite similar. This fully heterogeneous algorithm takes a relatively long time to converge with 200 episodes not yet leading to convergence. In the early stage of training, the Colight algorithm showed a fast convergence speed, which indicated that the algorithm could effectively use the information of the traffic network and quickly find a more reasonable traffic signal control policy, but it could not learn the optimal policy. Compared with other three methods, HAMH-PPO performs well in terms of convergence speed and final learning results. Therefore, we conclude that the HAMH-PPO model can learn better policy methods and have excellent convergence speed.

\subsection{Effectiveness of HAMH-PPO (RQ2)}

\begin{figure}[!h]
    \centering
    \subfigure[Hyper-actions (with a dimension of two) of 12 intersections in the dataset $D_{jinan1}$ at two times, 1400s and 3500s.]{
    \label{fig:RQ2_1}
    \centering
    \includegraphics[width=13.5cm]{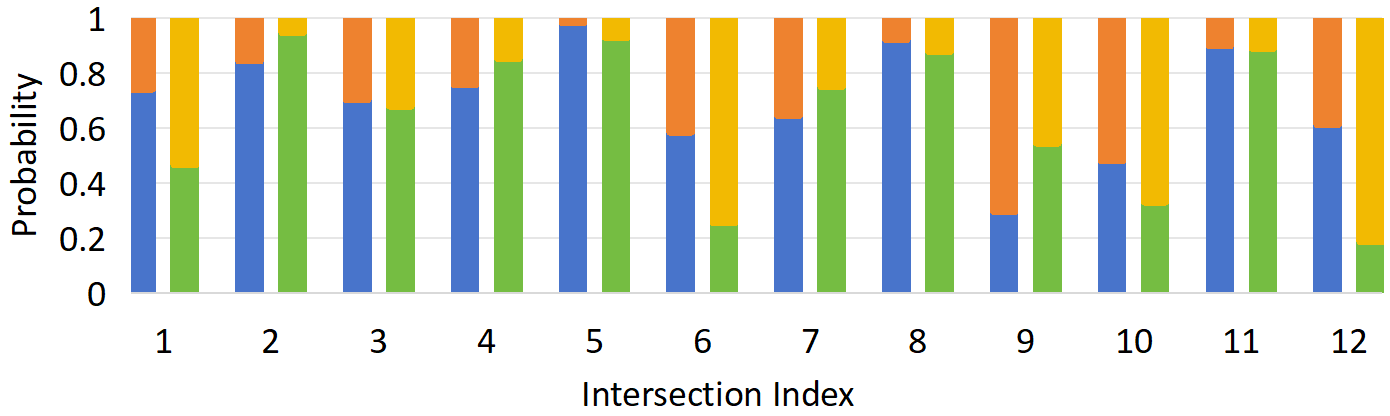}}
    \subfigure[The traffic flow statistics for intersections 1 and 11.]{
    \label{fig:RQ2_2}
    \includegraphics[width=6.5cm]{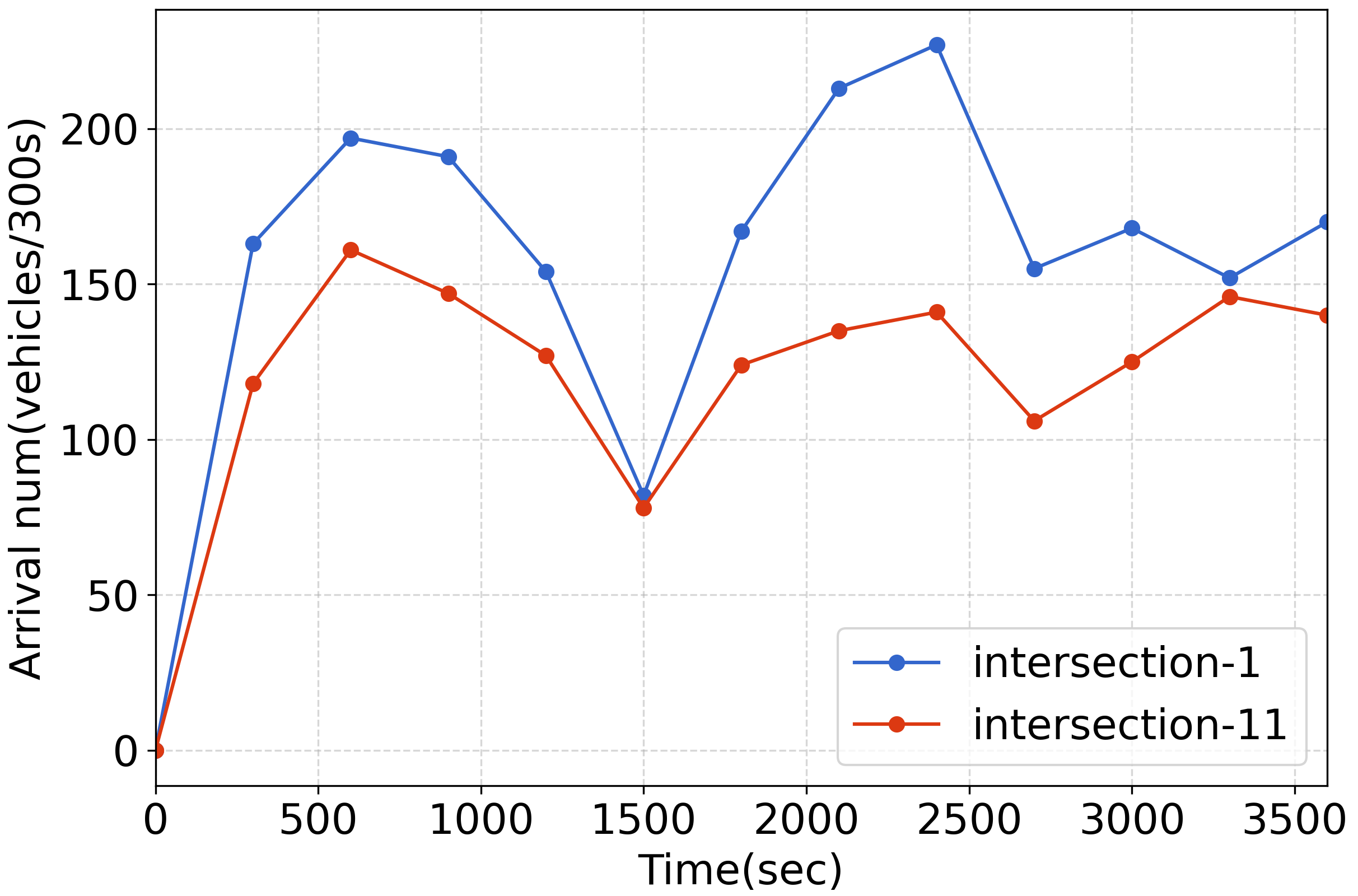}}
    \subfigure[The probability dynamics of the hyper-action for intersections 1 and 11.]{
    \label{fig:RQ2_3}
    \includegraphics[width=6.5cm]{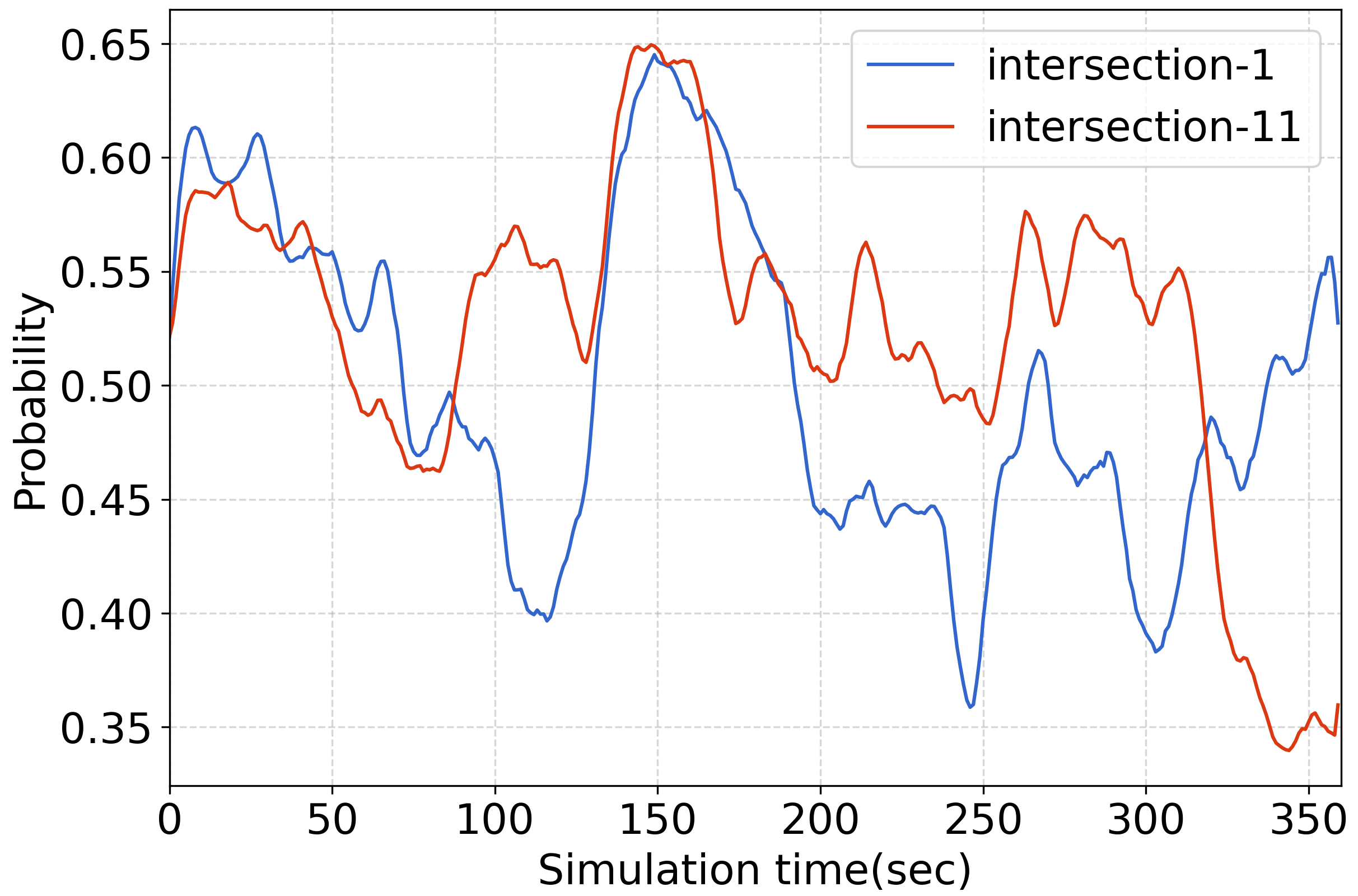}}    
    \caption{Effectiveness of hyper-action in HAMH-PPO (dataset:$D_{Jinan1}$).}
    \label{fig:enter-label}
\end{figure}

In this section, we will visualize the intersections' diversity value estimates to demonstrate this method's feasibility. We conducted experiments on the real road network of 12 intersections in $D_{Jinan1}$ to facilitate graphical representation. The traffic flow distribution in $D_{Jinan1}$'s road network has a large variance, making the diversity of intersections more apparent. Fig.~\ref{fig:RQ2_1} shows the values of hyper-action for all 12 intersections at time $t=1400s$ and $t=3500s$, with a hyper-action dimension of 2. For example, in the first intersection, the two bar graphs represent the values of hyper-action at 1400s as $0.7414 \& 0.2586$, and at 3500s as $0.466 \& 0.534$. The importance of the value function varies at different time points, indicating that our algorithm can calculate preference-based value function estimates based on the environment of intersections.

To further demonstrate the variations and effects of hyper-action in continuous time, we conducted additional experiments. We counted traffic flows (vehicles/300s) at two intersections, \ie, $1$ and $11$. As shown in Fig~\ref{fig:RQ2_2}, the traffic flow of 
intersection $1$ has a higher traffic volume than $11$. Additionally, during the time intervals of 1200s to 1500s and 3000s to 3300s, traffic flows at the two intersections are highly similar. We still set the output dimension of hyper-action to 2 and calculate the probability values of hyper-action taken by the agent at each step after training for 200 episodes. In the simulator, actions are sampled every ten seconds, so the range of the horizontal axis is from 0 to 360. As shown in Fig~\ref{fig:RQ2_3}, we only plot the values of the first dimension of the hyper-action. It can be seen that the probability values at the two intersections are extremely similar at the 150th step (\ie, the 1500s), but the probabilities change opposite after the 330th step (\ie, 3300s), which is consistent with the distribution of traffic flow at the two intersections. This indicates that the hyper-action network can dynamically adjust the importance of the value function based on different environments, thereby taking into account the preferences between intersections.

\subsection{Ablation Study (RQ3)}
\begin{figure}[t]
    \centering
    \subfigure[Ablation study in the dataset $D_{Jinan1}$.]{
    \label{fig:RQ3_2}
    \includegraphics[width=6.5cm]{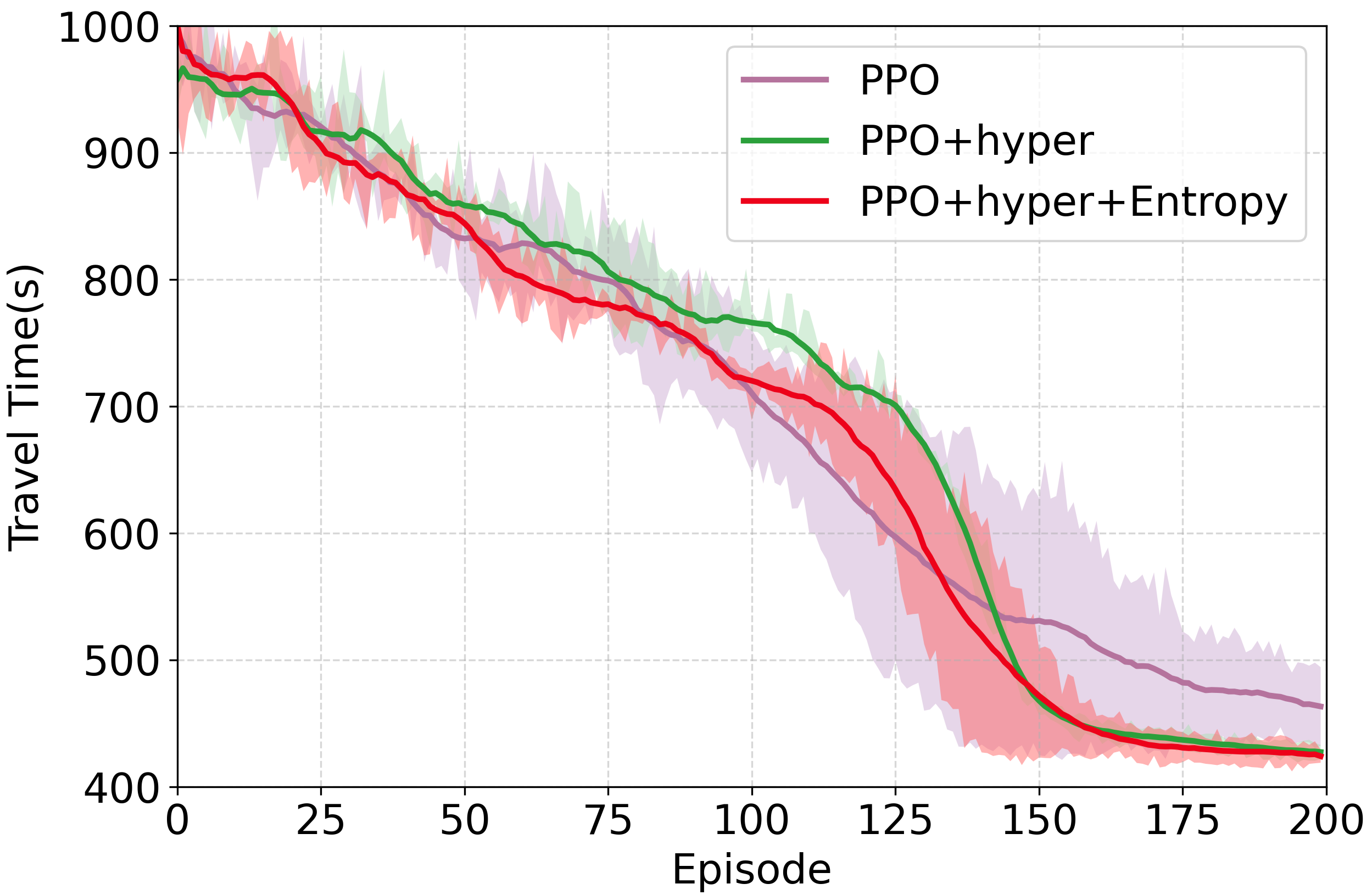}} 
    \subfigure[Ablation study in the dataset $Grid_{4\times4}$. ]{
    \label{fig:RQ3_1}
    \includegraphics[width=6.5cm]{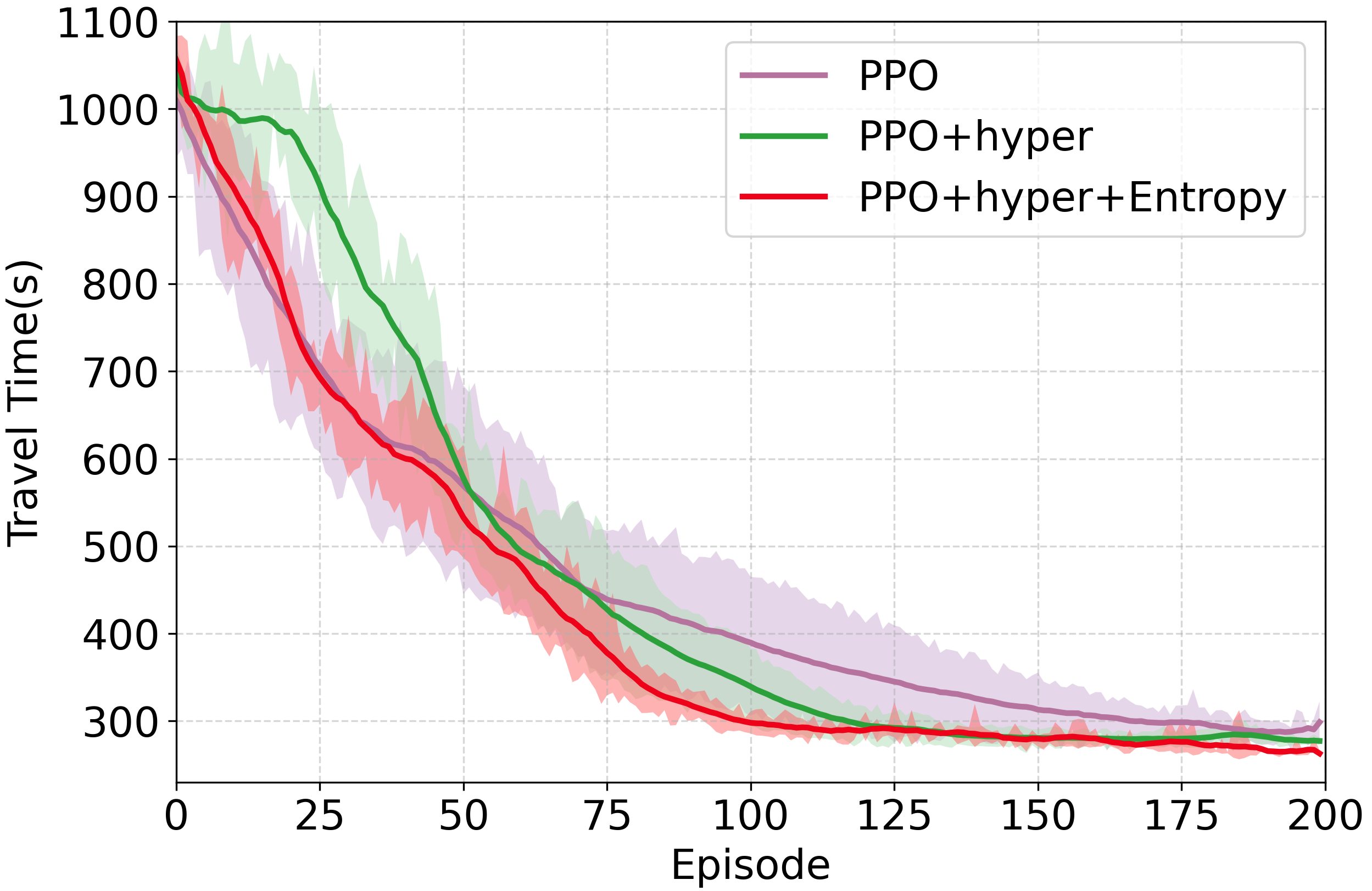}}   
    \caption{Learning dynamics of PPO+hyper+Entrony, PPO+hyper and PPO in two traffic flow distributions.}
    \label{fig:Ablation-1}
\end{figure}

\begin{figure}[h]
    \centering
    \subfigure[The comparison of different value numbers in critic networks after training 200 episodes in datasets $D_{Jinan1}$, $D_{Jinan2}$, and $Grid_{4\times4}$.]{
    \label{fig:Ablation-2}
    \includegraphics[width=6.4cm]{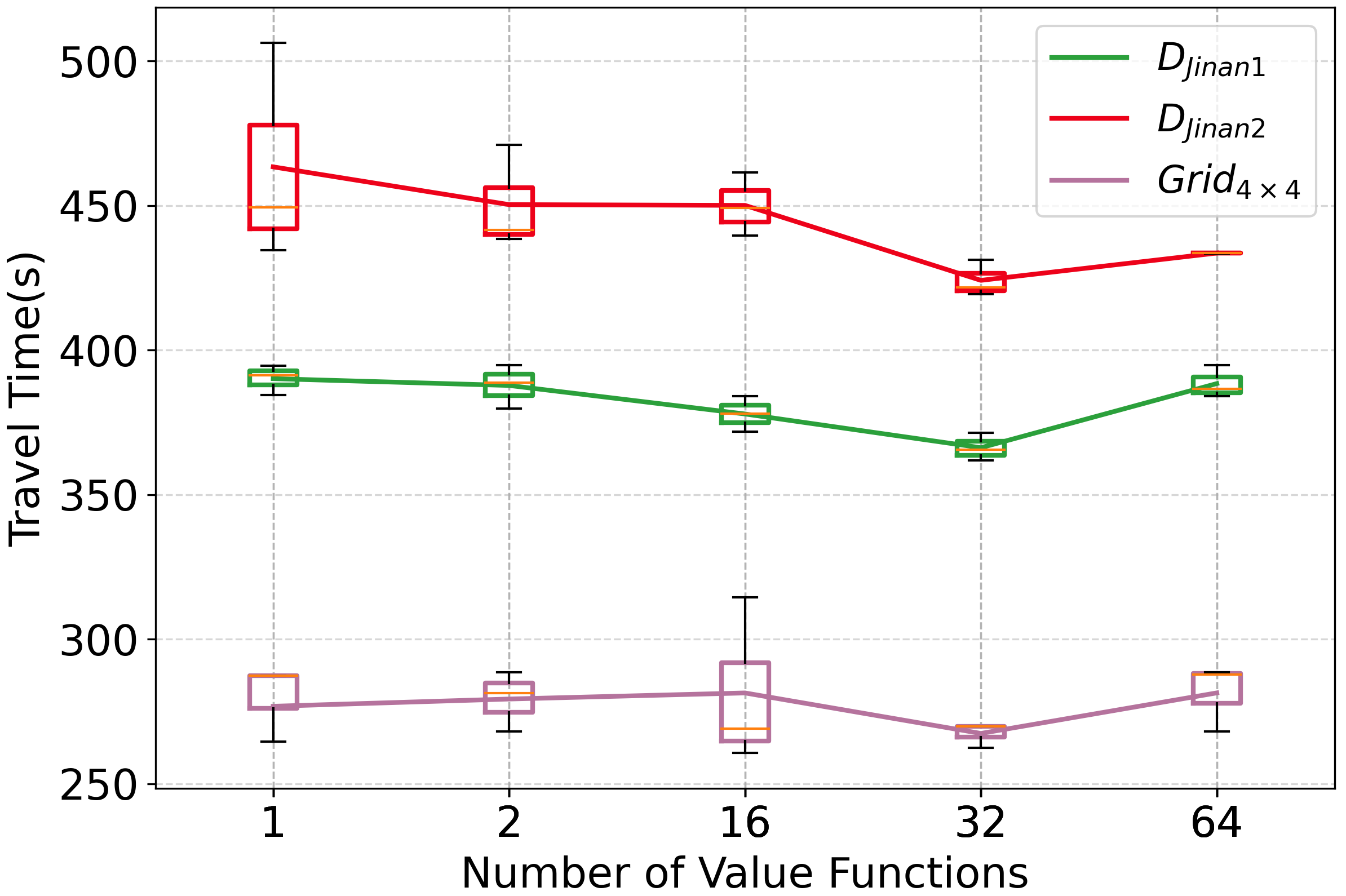}}
    \hfill
    \subfigure[The results of RL methods under four scales in the dataset $Grid_{4\times4}$. (
    {{\fontsize{8}{10}\selectfont HAMH-PPO} \raisebox{1pt}[1pt][1pt]{\colorbox[HTML]{5B9BD5}{}} },
    {{\fontsize{8}{10}\selectfont Colight} \raisebox{1pt}[1pt][1pt]{\colorbox[HTML]{ED7D31} {}} },
    {{\fontsize{8}{10}\selectfont PNC-HDQN} \raisebox{1pt}[1pt][1pt]{\colorbox[HTML]{A5A5A5} {}} },
    {{\fontsize{8}{10}\selectfont MA2C} \raisebox{1pt}[1pt][1pt]{\colorbox[HTML]{FFC000} {}} },
    {{\fontsize{8}{10}\selectfont MPlight} \raisebox{1pt}[1pt][1pt]{\colorbox[HTML]{4472C4} {}} }).
    ]{
    \label{fig:Ablation-3}
    \includegraphics[width=6.5cm]{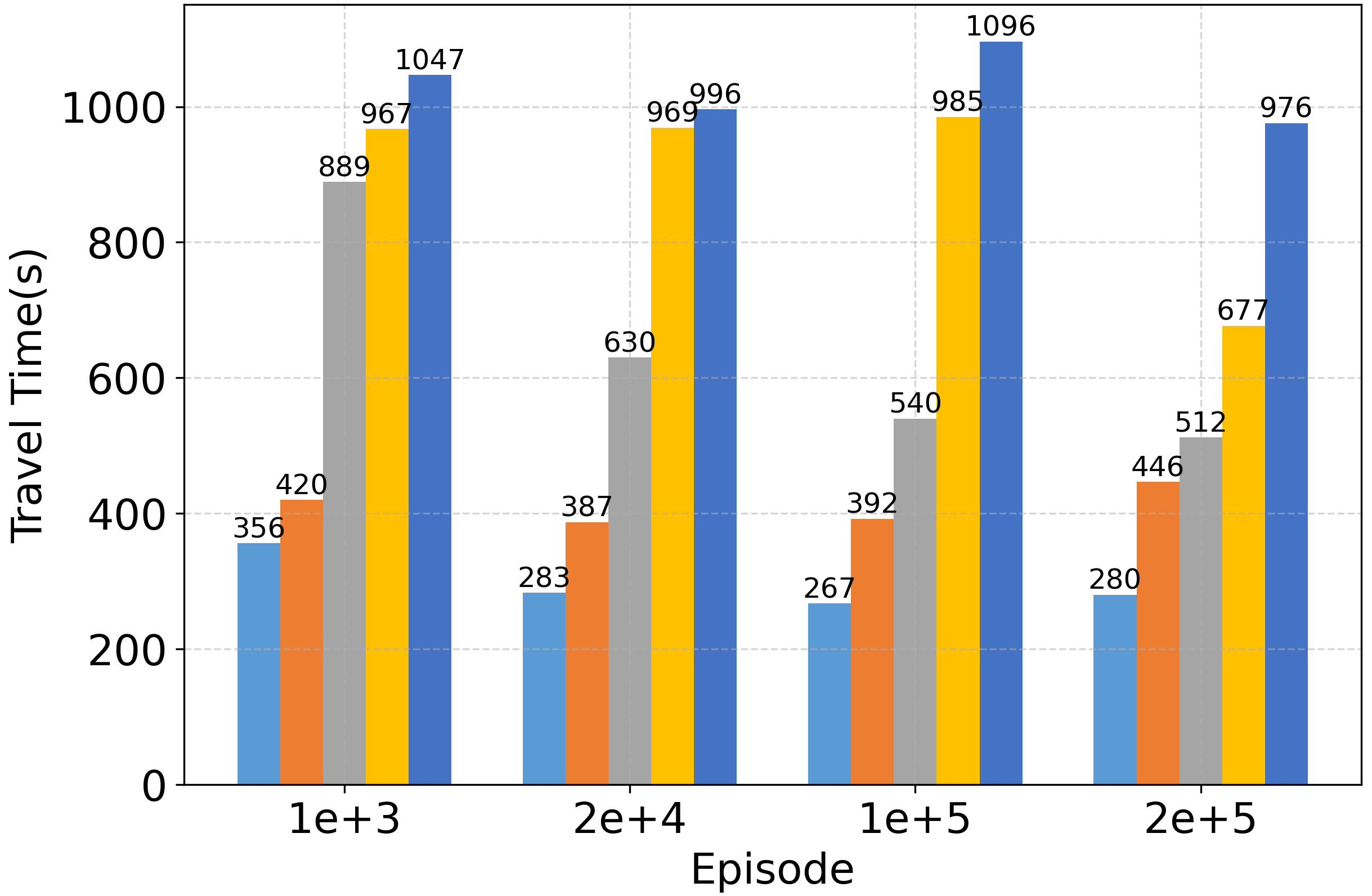}}
    \caption{Experimental comparisons of algorithms in the dimensions of hyper-action and neural network parameter scale.}
    \label{fig:RQ3-2}
\end{figure}
\subsubsection{The effectiveness of the components}
The hyper-action proposed in this paper is used to capture the preferences of intersections, while entropy is employed to ensure more even updates of the value function, to facilitate effective learning for all value functions. To validate the effectiveness of the hyper-action and weight entropy, we conducted ablation experiments with all methods sharing agent parameters. Our algorithm is \textit{PPO+hyper+Entrony}, where \textit{PPO+hyper} contains only hyper-action network. From Fig.~\ref{fig:Ablation-1}, it can be observed that adding hyper-action network and weight entropy produces the best performance and has a faster rate of convergence.

\subsubsection{The effectiveness of the dimensions in hyper-action}
To further verify the influence of the hyper-action dimension, we set the value functions in different dimensions to compare the training results. Considering the complexity of the collaborative scenarios is directly influenced by traffic density, we selected datasets 1 and 2 for the Jinan road network, which feature heavy traffic flows. In Fig.~\ref{fig:Ablation-2}, we calculate the travel time when different value function counts converged for three traffic flow distributions in two traffic environments. The line graph represents the average of three training results, and the orange line in the box denotes the median. The experimental results indicate that the dimension of hyper-action (or size of candidate values in the multi-head critic) has the best effect around 32. Too few candidate values prevent HAMH-PPO from fully representing the individual characteristics of each intersection, resulting in inaccurate value estimation. Conversely, the features of the intersection are excessively represented with too many value functions, resulting in interference.

\subsubsection{Comparison of neural network parameter scale}
Our algorithm combines the advantages of shared and non-shared parameter algorithms.  Parameter sharing can improve model training efficiency but cannot consider the preferences of each intersection. Non-parameter sharing trains a set of parameters for each agent, and the parameter scale will be very large when there are many intersections in the environment. Our algorithm retains the advantages of parameter sharing and discards its disadvantages, to achieve good results even with a smaller scale of neuron network. We compared the performance of five reinforcement learning-based methods at four different parameter scales. As shown in Fig.~\ref{fig:Ablation-3}, our algorithm performs the best under all four scales of networks. Especially when the parameter scale is 1000, HAMH-PPO has achieved the effect that other algorithms cannot achieve. Additionally, it can be observed that when the parameter scale is small, all algorithms fail to explore the best policy due to the neural network's inability to represent the features of the intersection fully. As the number of neural network parameters increases, the performance of shared parameter algorithms (HAMH-PPO, Colight, MPlight) initially improves and then tends to be stable. The performance of HAMH-PPO decreases when the number of parameters is very large, which we consider to be a problem of overfitting.  Increasing the weight entropy coefficient can appropriately alleviate the overfitting of the neural network.

\section{Conlcusion}
\label{sec:Conlcusion}
The Hyper-Action Multi-Head Proximal Policy Optimization (HAMH-PPO) algorithm proposed in this paper offers an effective solution for traffic signal control in large-scale intersection scenarios. By introducing the concepts of hyper-action and multi-value estimation, HAMH-PPO achieves personalized policy learning for intersections with non-independent and identically distributed (non-iid) observational distributions, while maintaining efficient parameter sharing. Experimental results demonstrate that HAMH-PPO enhances traffic efficiency while ensuring efficient training. Leveraging hyper-action from the HA-Actor, HAMH-PPO can learn preferential policies without increasing the number of network parameters, thus better accommodating the specific needs of different intersections. Furthermore, the critic network integrated with multi-value estimation provides a more accurate assessment of the average travel time for each intersection, further improving the performance of the policies. By efficiently sharing parameters and learning personalized policies, the number of parameters is significantly reduced thus lowering deployment costs and enhancing the feasibility and efficiency of the algorithm in practical applications.

In the future, we will continue to study how to further enrich the information encompassed in hyper-action, which can take the form of communication with other intersections or predictions of the overall traffic state. The goal is to provide a more comprehensive and detailed description of traffic conditions so that more efficient cooperation among agents can be achieved.

\section*{Acknowledgments}
This work was supported by the National Natural Science Foundation of China under Grant 62376048, the Shandong Provincial Natural Science Foundation, China under Grant ZR2022LZH002, and the Fundamental Research Funds for the Central Universities under Grant 3132024250.

\bibliographystyle{elsarticle-num-names}
\bibliography{main}

\end{document}